\pdfoutput=1
\RequirePackage{fix-cm}
\PassOptionsToPackage{table}{xcolor}
\PassOptionsToPackage{headheight=15.5mm,headsep=3mm}{geometry}
\documentclass[letterpaper]{bytedance_seed}

\usepackage{amsmath,amssymb}
\usepackage{array}
\usepackage{mathtools}
\usepackage{amsthm}
\usepackage{algorithm}
\usepackage{algorithmic}
\usepackage[shortlabels,inline]{enumitem}
\usepackage{xspace}
\usepackage[colorlinks,allcolors=seedblue]{hyperref}
\usepackage[capitalize,noabbrev,nameinlink]{cleveref}

\DeclareFontShape{T1}{cmr}{b}{it}{<->ssub*cmr/bx/it}{}
\AtBeginEnvironment{tabular}{\renewcommand{\textbf}[1]{{\bfseries #1}}}


\definecolor{oursrow}{HTML}{EEEAFB}
\definecolor{oursbg}{HTML}{EEEAFB}
\definecolor{tabgray}{HTML}{52514E}
\definecolor{notespurple}{HTML}{6A4FD0}
\definecolor{videoorange}{HTML}{EB6834}
\definecolor{posgreen}{HTML}{1B7837}

\newtcolorbox{findingbox}{colback=oursrow,colframe=notespurple!70,boxrule=0.5pt,arc=1.5pt,
  left=3pt,right=3pt,top=1.5pt,bottom=1.5pt,before skip=3.5pt,after skip=3.5pt}
\newcounter{finding}
\newcommand{\finding}[2]{\refstepcounter{finding}%
  \begin{findingbox}\small\textbf{Finding~\thefinding. #1} #2\end{findingbox}}

\theoremstyle{plain}
\newtheorem{theorem}{Theorem}[section]
\newtheorem{proposition}[theorem]{Proposition}

\theoremstyle{definition}

\crefname{assumption}{Assumption}{Assumptions}
\Crefname{assumption}{Assumption}{Assumptions}
\crefname{proposition}{Proposition}{Propositions}
\Crefname{proposition}{Proposition}{Propositions}
\AddToHook{env/proposition/begin}{\crefalias{theorem}{proposition}}

\newcommand{\E}{\mathbb{E}}

\newcommand{\KL}{\mathbb{D}_{\mathrm{KL}}}
\newcommand{\Ind}{\mathbb{I}}

\DeclareMathOperator{\clip}{clip}

\newcommand{\norm}[1]{\left\|#1\right\|}
\newcommand{\abs}[1]{\left|#1\right|}

\newcommand{\method}{ActiveForager\xspace}
\newcommand{\rlmethod}{TUPO\xspace}

\newcommand{\Vpref}{\mathcal{V}_{\le t}}
\newcommand{\Wrec}{\mathcal{W}_t^{\mathrm{rec}}}
\newcommand{\Ahist}{\mathcal{A}_{<t-w+1}}
\newcommand{\Irec}{\mathcal{I}_{\mathrm{rec}}}
\newcommand{\Iforage}{\mathcal{I}_{\mathrm{for}}}
\newcommand{\Ians}{\mathcal{I}_{\mathrm{ans}}}
\newcommand{\MSDAM}{\mathbf{M}^{\mathrm{SDAM}}}
\newcommand{\DeltaP}{\Delta P}
\newcommand{\DeltaM}{\Delta M}

\newcommand{\DeltaI}{\Delta \mathcal{I}}
\newcommand{\RIG}{R_{\mathrm{IG}}}
\newcommand{\Racc}{R_{\mathrm{acc}}}
\newcommand{\Ranchor}{R_{\mathrm{fmt}}}
\newcommand{\VCCV}{V_{\mathrm{act}}}
\newcommand{\piref}{\pi_{\mathrm{ref}}}
\newcommand{\pitheta}{\pi_{\theta}}

\newcommand{\ActAnswer}{Answer\xspace}
\newcommand{\ActZoom}{ZoomIn\xspace}
\newcommand{\ActRecall}{Recall\xspace}
\newcommand{\ActWait}{Wait\xspace}
\newcommand{\EvCard}{EvidenceCard\xspace}

\newlength{\logoheight}
\fancypagestyle{firststyle}{%
  \fancyhf{}%
  \fancyhead[L]{%
    \includegraphics[height=\logoheight]{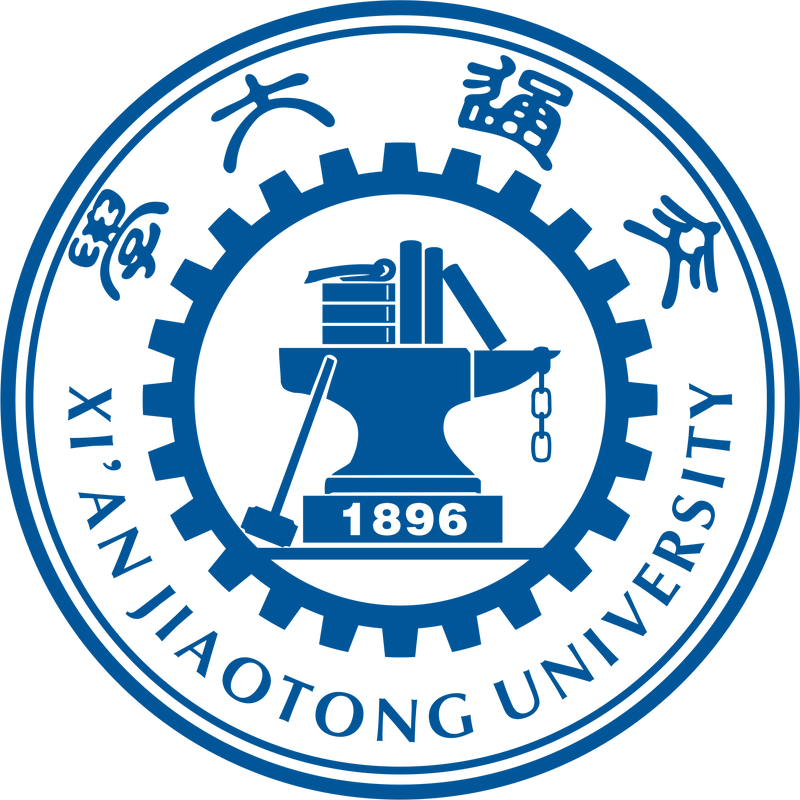}\hspace{4mm}%
    \includegraphics[height=\logoheight]{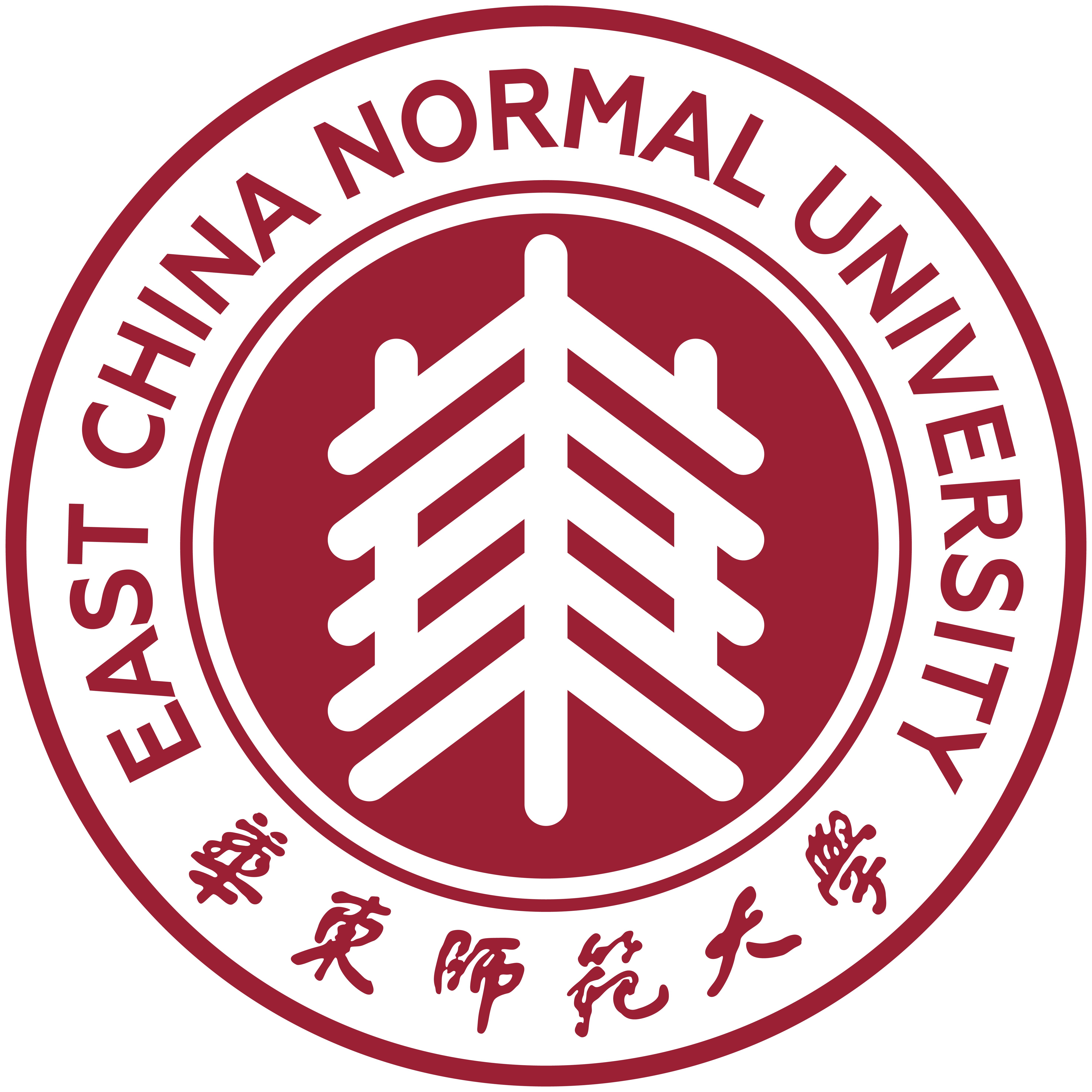}\hspace{4mm}%
    \includegraphics[height=\logoheight]{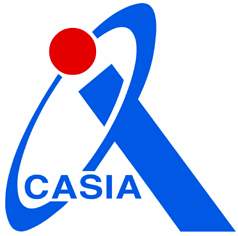}\hspace{4mm}%
  }%
  \fancyfoot[C]{\thepage}%
}

\title{Look Closer, Look Back, or Wait\\for Agentic Streaming Video Understanding}
\hypersetup{
  pdftitle={Look Closer, Look Back, or Wait for Agentic Streaming Video Understanding},
  pdfauthor={Pengyu Li, Zhitao Gao, Lingling Zhang, Wenxuan Huang, YiCheng Xiao, Chen Zheng, Jun Liu}
}

\author[1]{Pengyu Li}
\author[1]{Zhitao Gao}
\author[1,\dagger]{Lingling Zhang}
\author[2]{Wenxuan Huang}
\author[3]{YiCheng Xiao}
\author[4]{Chen Zheng}
\author[1]{Jun Liu}

\affiliation[1]{Xi'an Jiaotong University}
\affiliation[2]{East China Normal University}
\affiliation[3]{Institute of Automation, Chinese Academy of Sciences}
\affiliation[4]{Open University of China}

\contribution[\dagger]{Corresponding author}
\abstract{Streaming video models must answer questions about a live video under memory and latency limits. A short recent window supports current-scene perception but loses past events. Adding historical frames restores context but can divert attention from the present scene and induce hallucinations. Historical retrieval also cannot recover missing spatial detail or reveal clues that have not yet appeared. We introduce ActiveForager, a streaming video agent that keeps a clean four-frame working window and acquires additional evidence on demand. For each query, it answers directly, inspects recent frames at higher resolution, recalls past events, or waits for new frames. During recall, the same VLM verifies candidate clips in a separate context and returns a compact evidence card. Stream-Decoupled Attention Masking (SDAM) isolates the visual streams during synthesis. Tool-Utility Policy Optimization (\rlmethod) rewards tool calls when their evidence increases the likelihood of the correct answer. Across Qwen2.5-VL-7B and Qwen3-VL-8B, ActiveForager improves present perception, historical recall, and proactive response together. It reaches 71.3\% overall accuracy on OVO-Bench with 64\% fewer main-context visual tokens than prior cascades. The learned policy also transfers to StreamingBench and multi-hop video reasoning on Video-Holmes.
}
\checkdata[Contact]{\email{lipengyu.tiez@stu.xjtu.edu.cn}}
\checkdata[Code]{\url{https://github.com/TiezMind/ActiveForager}}

\begin{document}
\maketitle

\begin{figure}[t]
  \centering
  \includegraphics[width=\textwidth]{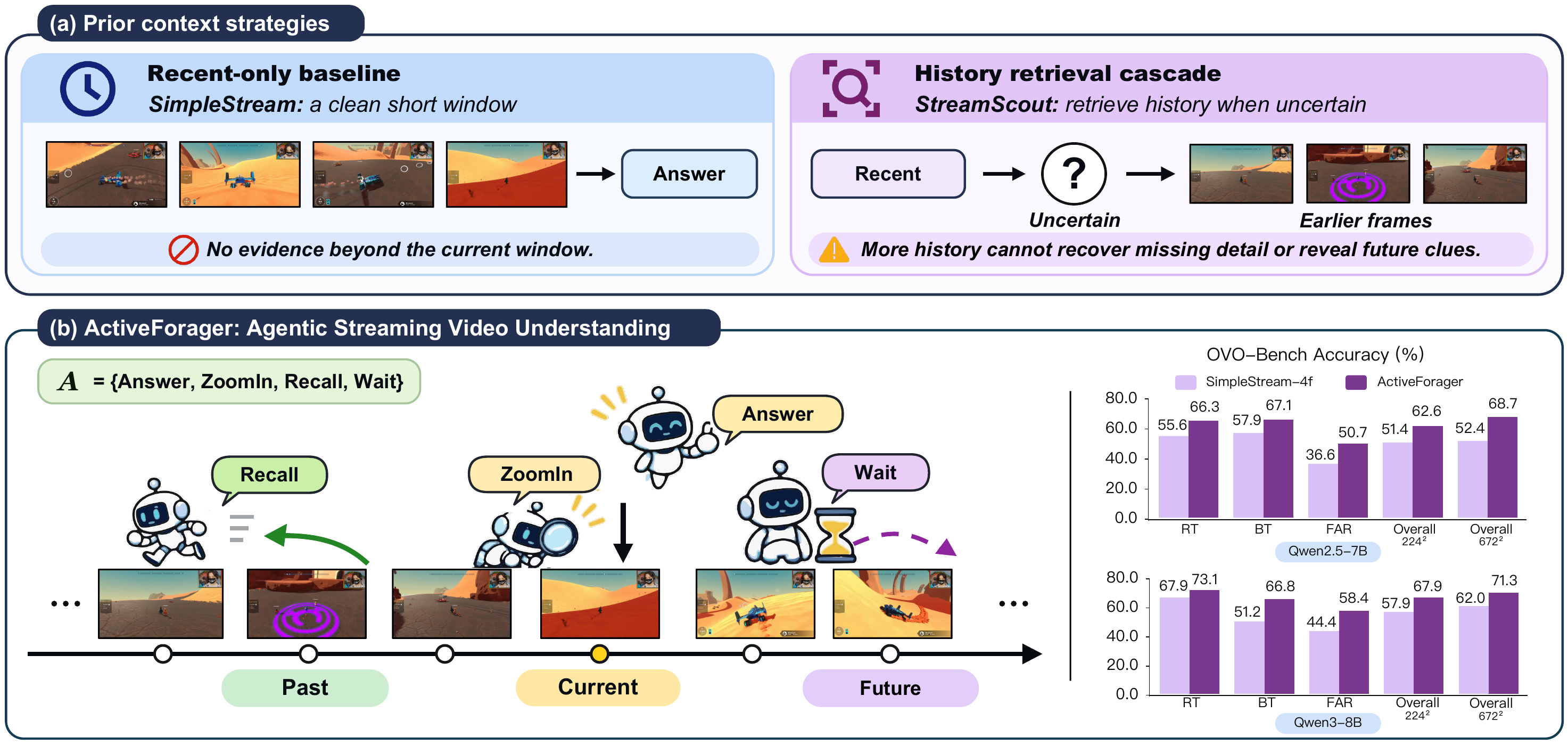}
  \caption{\textbf{Selective evidence acquisition for streaming video understanding.} (a) Recent-only windows lose past context. Retrieving history does not address missing current-scene detail or clues that have yet to appear. (b) \method answers directly, inspects recent frames, recalls past events, or waits for new evidence. Insets compare overall OVO-Bench accuracy of SimpleStream-4f on both backbones and both resolutions. RT, BT and FAR denote real-time perception, backward tracing and proactive response at $224^2$.}
  \label{fig:pareto_frontier}
\end{figure}

\section{Introduction}
\label{sec:intro}

Streaming vision-language models (VLMs) answer questions about live video under memory and latency limits \citep{chen2024videollmonline,niu2025ovobench,lin2024streamingbench}. At each query timestamp, the model must perceive the current scene, recall earlier events, and wait when a decisive clue has not yet appeared \citep{niu2025ovobench,liu2026thinkstream}. These demands must be met using the video observed so far.

A short recent window supports current-scene perception, as demonstrated by the strong SimpleStream-4f baseline \citep{shen2026simplestream}. Questions about earlier events, however, may require evidence outside this window. Memory-based models preserve historical information to support such queries \citep{zhang2024flashvstream,li2025hermes}. When historical visual tokens share the answering context with recent frames, they can divert attention from the present scene. The challenge is therefore to recover useful history while protecting current-scene perception.

Recent cascades such as OASIS and StreamScout make access to history selective \citep{liang2026oasis,zhang2026streamscout}. They answer from recent frames first and retrieve history when uncertain. This helps when the missing evidence lies in the past. A query about small text, however, may need a closer view of the current scene. A query about an ongoing action may need more observation before a decisive clue appears. A streaming agent must therefore identify what evidence is missing and choose how to acquire it.

We introduce \textbf{\method}, a streaming video agent that chooses how to acquire evidence for each query (\cref{fig:pareto_frontier}). It keeps a clean recent working window and selects among direct answering (\ActAnswer), closer inspection (\ActZoom), historical retrieval (\ActRecall), and waiting (\ActWait). During recall, the same VLM checks candidate clips in a separate context for evidence of the queried event. It returns an \EvCard{} with a factual note and at most one anchor frame. A negative verification result adds no historical visual tokens to the main context. Stream-Decoupled Attention Masking (SDAM) controls attention between the recent window and the additional visual evidence during synthesis.

Learning when to use tools requires a signal beyond answer correctness. Pretrained backbones differ in how often they invoke each tool. Standard cost-penalized GRPO can suppress tool use because coarse frames often suffice for multiple-choice guessing. Rewards that favor historical retrieval can also leave \ActZoom{} and \ActWait{} with little useful training signal \citep{yang2026paravt,zhang2026streamscout}. We address these issues with \textbf{Tool-Utility Policy Optimization (\rlmethod)}, a GRPO-based training scheme. It exposes the agent to contexts with missing spatial detail, missing history, or incomplete actions. A tool earns a positive utility reward when its evidence increases the likelihood of the correct answer. Across OVO-Bench, StreamingBench, and Video-Holmes \citep{cheng2025videoholmes}, \method improves perception, recall, and proactive response. It reaches 71.3\% overall accuracy on OVO-Bench with 64\% fewer main-context visual tokens than StreamScout-R.

\noindent We summarize our main contributions as follows.
\begin{itemize}
\item We introduce \method, a streaming video agent that separates its recent working window from CPU evidence storage. It can answer directly (\ActAnswer), look closer (\ActZoom), look back (\ActRecall), or wait (\ActWait) (\cref{sec:method_buffer}).
\item We combine isolated evidence verification (\EvCard) with Stream-Decoupled Attention Masking (SDAM) to fuse relevant evidence and curb false-premise hallucinations (\cref{sec:method_sdam}).
\item We propose \rlmethod, which learns selective tool use through counterfactual utility rewards. The trained agent sustains all three foraging tools and transfers across streaming and multi-hop video benchmarks (\cref{sec:method_rl,sec:experiments}).
\end{itemize}

\section{Related Work}
\label{sec:related_work}

\paragraph{Streaming video understanding and memory compression.}
Online video models process incoming streams under bounded latency and memory \citep{chen2024videollmonline,niu2025ovobench,lin2024streamingbench}. VideoLLM-online \citep{chen2024videollmonline} and LiveCC \citep{chen2025livecc} study dialogue timing. Memory-augmented models retain information across longer video histories. Flash-VStream clusters spatial-temporal features \citep{zhang2024flashvstream}, and Dispider separates perception from decision making \citep{qian2025dispider}. TimeChat-Online prunes redundant tokens \citep{yao2025timechatonline}. StreamForest builds an event memory forest \citep{zeng2025streamforest}, while HERMES combines episodic and recency KV caches \citep{li2025hermes}. SimpleStream shows that a short recent window can outperform these more elaborate memory systems \citep{shen2026simplestream}. \method keeps this initial window clean and acquires additional evidence on demand.

\begin{figure*}[t]
    \centering
    \includegraphics[width=\textwidth]{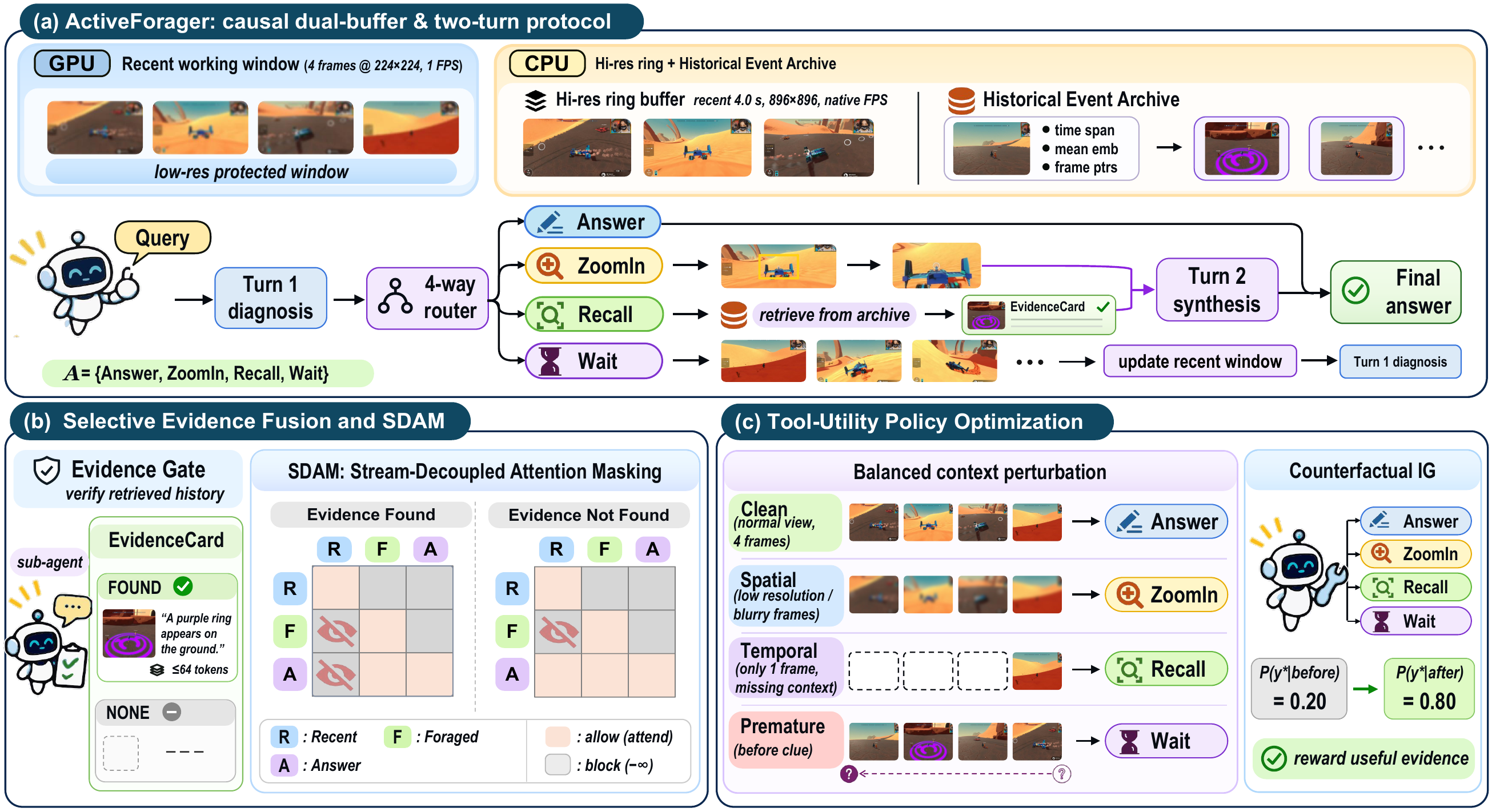}
    \caption{\textbf{\method architecture and \rlmethod training.} (a) A compact GPU window and CPU evidence storage support four-way routing. \ActAnswer{} returns directly. \ActWait{} advances the stream before routing resumes. (b) A separate context verifies recalled history. \emph{Found} returns a factual note and one anchor frame. \emph{None} adds no historical visual tokens to the main agent. SDAM controls attention to recent and acquired evidence during synthesis. (c) \rlmethod combines balanced context perturbations with counterfactual tool-utility rewards. The probability change illustrates the utility signal.}
    \label{fig:architecture}
\end{figure*}

\paragraph{Agentic video reasoning and temporal cascades.}
Agentic frameworks choose when and where to inspect visual evidence \citep{guo2025framethinker,chen2026egagent,zeng2026videoo3}. FrameThinker \citep{guo2025framethinker} and EGAgent \citep{chen2026egagent} retrieve segments for offline video reasoning. Video-o3 applies task-decoupled attention masking to offline tool trajectories \citep{zeng2026videoo3}. For live video, OASIS \citep{liang2026oasis} and StreamScout \citep{zhang2026streamscout} retrieve history after an uncertain initial answer. ThinkStream studies proactive response timing \citep{liu2026thinkstream}. Historical retrieval does not provide missing current-scene detail or reveal upcoming clues. \method supports inspection, recall, and waiting within one routing policy. An isolated sub-agent verifies recalled evidence before SDAM controls its use during synthesis.

\paragraph{Reinforcement learning for multimodal tool use.}
GRPO provides a basis for training multimodal tool use \citep{feng2025videor1,liu2025drgrpo,li2026visual,gao2026aero,li2026agtao,li2026thinkingvlmsneedlooking}. ParaVT identifies the \emph{Tool Prior Paradox}, where tool costs favor direct guessing over evidence acquisition \citep{yang2026paravt}. EVA studies related guessing shortcuts \citep{wang2026eva}. Dr.\ GRPO removes group standard-deviation normalization \citep{liu2025drgrpo}, and TurnWise balances gradients across tool calls and answers \citep{zhou2026turnwise}. A reward focused on historical retrieval can make \ActRecall{} dominate training while other tools stop being selected. \rlmethod combines balanced training contexts with counterfactual utility rewards for \ActZoom{}, \ActRecall{}, and \ActWait{}.

\section{Preliminaries}
\label{sec:preliminaries}

\subsection{Causal Budgeted Streaming Formulation}
\label{sec:prelim_formulation}

When query $q_t$ arrives at time $t$, the agent can observe only the video prefix $\Vpref = \{f_\tau \mid \tau \le t\}$. Its main visual context has a bounded token budget. Following \citet{shen2026simplestream} and \citet{niu2025ovobench}, we evaluate Real-Time Visual Perception ($\mathrm{RT}$), Backward Tracing ($\mathrm{BT}$), and Forward Active Responding ($\mathrm{FAR}$). SimpleStream-4f uses the four most recent frames $\Wrec$ as the reference point. We define perception gain as $\DeltaP(m) = \mathrm{RT}(m) - \mathrm{RT}(\text{SimpleStream-4f})$ and memory gain as $\DeltaM(m) = \mathrm{BT}(m) - \mathrm{BT}(\text{SimpleStream-4f})$. The goal is to improve both gains while maintaining strong proactive response.

\subsection{Softmax Attention Dilution and Tool Collapse}
\label{sec:prelim_dilution}

Historical and recent visual tokens compete for attention in a shared context. Consider a query token that attends to $L_{\mathrm{rec}}$ recent visual tokens $\Irec$. Let $Z_{\mathrm{rec}}$ be their Softmax partition sum and $\alpha_0(\Irec)$ their baseline attention mass. Appending $L_{\mathrm{ret}}$ historical tokens $\Iforage$ introduces an additional partition sum $Z_{\mathrm{ret}}$. Their relative per-token affinity is $\bar{\rho} = (L_{\mathrm{rec}} Z_{\mathrm{ret}}) / (L_{\mathrm{ret}} Z_{\mathrm{rec}})$. Under unmasked causal attention, the recent-scene attention mass becomes
\begin{equation}
\label{eq:dilution_bound}
\alpha_{\mathrm{naive}}(\Irec) = \frac{\alpha_0(\Irec)}{1 + \alpha_0(\Irec) \frac{L_{\mathrm{ret}}}{L_{\mathrm{rec}}} \bar{\rho}} < \alpha_0(\Irec).
\end{equation}
The attention mass on recent frames decreases as additional historical tokens enter the unmasked context (\cref{app:proof_dilution}). This motivates a clean initial window, evidence gating after verification, and controlled attention during synthesis.

\paragraph{Why tool use collapses.}
Training the policy over $\{\ActAnswer, \ActZoom, \ActRecall, \ActWait\}$ exposes two forms of reward imbalance (\cref{app:proof_crowding}). Tool costs can make direct guesses more rewarding when coarse frames suffice on many prompts. Group-centered updates then suppress evidence acquisition. History-only RL methods introduce a different imbalance \citep{yang2026paravt,zhang2026streamscout}. Historical retrieval receives most of the positive reward contrast, and \ActZoom{} or \ActWait{} can stop being selected. Balanced contexts and tool-specific utility measurements provide a learning signal for all three tools.

\section{Method}
\label{sec:method}

\method keeps a recent working window and stores additional visual evidence on the CPU. A routing policy chooses whether to answer, inspect recent frames, recall history, or wait (\cref{sec:method_buffer}). Historical evidence is verified in a separate context before SDAM controls its use during synthesis (\cref{sec:method_sdam}). \rlmethod trains the policy to select tools whose evidence improves the answer (\cref{sec:method_rl}).

\subsection{Working Memory and Four-Way Tool Routing}
\label{sec:method_buffer}

\paragraph{Recent working window ($\Wrec$).}
The GPU working context retains the four most recent frames at $224\times224$. After patch merging \citep{bai2025qwen25vl,yang2025qwen3vl}, this window occupies 255.5 visual tokens on average. A compact header records the current stream time and active frame count. Historical frames remain outside the initial working context.

\paragraph{CPU evidence storage and history archive.}
A CPU ring buffer stores high-resolution recent frames for \ActZoom{}. Frames outside the GPU working window are indexed in a history archive $\Ahist$. We group them into segments using a frame-change score computed from lightweight visual descriptors. Each segment records its time span, peak change score, and frame pointers. Buffer settings and segmentation details are given in \cref{app:implementation_details}.

\paragraph{Four-way tool routing.}
Given $(\Wrec, q_t)$, the policy $\pitheta$ produces a brief diagnostic rationale and selects $a_1 \in \{\ActAnswer, \ActZoom, \ActRecall, \ActWait\}$.
\begin{itemize}
    \item \textbf{Answer directly (\ActAnswer).} Return the final answer from the recent window without requesting additional visual evidence.
    \item \textbf{Inspect recent frames (\ActZoom).} Read a high-resolution crop or a dense frame burst from the CPU ring buffer. This helps resolve fine text, small attributes, or rapid motion (\cref{tab:supp_routing_ledger}).
    \item \textbf{Retrieve past events (\ActRecall).} Rank archived segments using temporal overlap and frame-change scores. Select candidates with separated midpoints and send representative frames for verification.
    \item \textbf{Wait for more evidence (\ActWait).} Advance the stream and reassess the query using newly observed frames. A bounded waiting budget limits repeated deferral. Clue-reveal evaluation also uses a separate readiness check to decide when to commit (\cref{app:wait_policy}).
\end{itemize}

\subsection{Selective Evidence Fusion and SDAM}
\label{sec:method_sdam}

When $a_1 \in \{\ActZoom, \ActRecall\}$, the agent obtains additional visual evidence. Appending it without attention control can dilute the recent window (\cref{eq:dilution_bound}). We therefore verify recalled evidence and control attention during synthesis.

\paragraph{Isolated evidence verification.}
\ActRecall{} reuses the same VLM weights in a separate context with a clean KV cache. This sub-agent inspects candidate frames and returns a structured \EvCard{}. Its status is $\mathrm{status} \in \{\text{Found}, \text{None}\}$. A \emph{Found} card contains a concise factual note and one anchor frame. For \emph{None}, the evidence gate passes no historical visual tokens to the main agent. This prevents unrelated retrieved frames from inducing false-premise hallucinations. During synthesis, the verified anchor or zoomed crop forms the additional visual evidence $\Iforage$. The resulting main context contains $L_{\mathrm{rec}} + L_{\mathrm{forage}}$ visual tokens.

\paragraph{Stream-Decoupled Attention Masking (SDAM).}
Vision delimiters identify the token spans for recent frames ($\Irec$), additional evidence ($\Iforage$), and answer tokens ($\Ians$). For query index $i$ and key index $j \le i$, we construct the additive mask $\MSDAM \in \{0,-\infty\}^{L\times L}$ as
\begin{equation}
\label{eq:sdam_mask}
\MSDAM_{i,j} = \begin{cases}
-\infty, & (i,j) \in \Iforage \times \Irec \cup \Irec \times \Iforage, \\
-\infty, & (i,j) \in \Ians \times \Irec \;\text{if Found}, \\
0, & \text{otherwise } (j \le i).
\end{cases}
\end{equation}
SDAM blocks direct attention between recent frames and additional visual evidence across all layers (\cref{fig:architecture}b). After \ActZoom{}, answer tokens can attend to both visual streams. When \ActRecall{} returns \emph{Found}, the verified historical anchor becomes the direct visual source for the answer. The earlier text context remains available.

\begin{table*}[t]
\caption{\textbf{Main comparison on OVO-Bench across passive and proactive streaming tracks.} \method advances accuracy across both resolution regimes while using at least 64\% fewer main-context visual tokens than prior cascades.}
\label{tab:ovobench_tracks}
\centering
\footnotesize
\renewcommand{\arraystretch}{0.95}
\setlength{\tabcolsep}{3.2pt}
\resizebox{\linewidth}{!}{%
\begin{tabular}{@{}ll r rrr rr@{}}
\toprule
 & & & \multicolumn{3}{c}{\textbf{Passive Streaming Tracks (9 Tasks)}} & \multicolumn{2}{c}{\textbf{Proactive \& All 12 Tasks}} \\
\cmidrule(lr){4-6} \cmidrule(l){7-8}
Method & Backbone & \textbf{Vis.\ Tokens} & \textbf{Real-Time (RT)} & \textbf{Backward (BT)} & \textbf{9-Avg} & \textbf{Forward (FAR)} & \textbf{Overall (12)} \\
\midrule
\multicolumn{8}{@{}l}{\emph{Prior Streaming Memory Systems}} \\
\quad Flash-VStream \citep{zhang2024flashvstream} & 7B & 1,024 & 28.37 & 27.38 & 28.04 & 45.09 & 32.30 \\
\quad Dispider \citep{qian2025dispider} & 7B & 1,460 & 54.55 & 36.06 & 48.39 & 34.72 & 44.97 \\
\quad StreamForest \citep{zeng2025streamforest} & Qwen2.5-7B & 1,024 & 61.20 & 52.01 & 58.14 & \textbf{53.49} & 56.98 \\
\quad HERMES$^\dagger$ \citep{li2025hermes} & Qwen2.5-7B & 4,096 & 68.98 & 49.42 & 62.46 & 41.10 & 57.12 \\
\midrule
\multicolumn{8}{@{}l}{\emph{High-Resolution Streaming Regime ($672^2$ Base Window)}} \\
\quad OASIS$^\dagger$ \citep{liang2026oasis} & Qwen2.5-7B & 10,000 & 68.06 & 53.18 & 63.10 & 40.45 & 57.44 \\
\quad StreamScout-R$^\dagger$ \citep{zhang2026streamscout} & Qwen2.5-7B & 6,559 & 76.50 & 69.55 & 74.18 & 40.25 & 65.70 \\
\quad SimpleStream-4f ($672^2$) \citep{shen2026simplestream} & Qwen2.5-7B & \textbf{2,299} & 62.80 & 47.46 & 57.69 & 36.60 & 52.41 \\
\rowcolor{oursrow}\quad \textbf{\method ($672^2$ Native Hi-Res)} & Qwen2.5-7B & 2,314 & \textbf{77.08} & \textbf{70.14} & \textbf{74.77} & \textbf{50.65} & \textbf{68.74} \\
\addlinespace[1.5pt]
\quad OASIS$^\dagger$ \citep{liang2026oasis} & Qwen3-8B & 10,000 & 77.36 & 59.14 & 71.29 & 47.80 & 65.41 \\
\quad StreamScout-R$^\dagger$ \citep{zhang2026streamscout} & Qwen3-8B & 6,559 & \textbf{79.84} & \textbf{75.52} & \textbf{78.40} & 45.96 & 70.29 \\
\quad SimpleStream-4f ($672^2$) \citep{shen2026simplestream} & Qwen3-8B & \textbf{2,299} & 75.67 & 52.25 & 67.86 & 44.38 & 61.99 \\
\rowcolor{oursrow}\quad \textbf{\method ($672^2$ Native Hi-Res)} & Qwen3-8B & 2,354 & 78.03 & 70.69 & 75.58 & \textbf{58.43} & \textbf{71.30} \\
\midrule
\multicolumn{8}{@{}l}{\emph{Budget-Constrained Streaming Regime ($224^2$ Base Window)}} \\
\quad SimpleStream-4f ($224^2$) \citep{shen2026simplestream} & Qwen2.5-7B & \textbf{255.5} & 55.60 & 57.86 & 56.35 & 36.60 & 51.41 \\
\rowcolor{oursrow}\quad \textbf{\method (Pretrained Agent)} & Qwen2.5-7B & 258.6 & 63.68 & 58.98 & 62.11 & 35.29 & 55.41 \\
\rowcolor{oursrow}\quad \textbf{\method (\rlmethod)} & Qwen2.5-7B & 295.1 & \textbf{66.27} & \textbf{67.11} & \textbf{66.55} & \textbf{50.65} & \textbf{62.57} \\
\addlinespace[1.5pt]
\quad SimpleStream-4f ($224^2$) \citep{shen2026simplestream} & Qwen3-8B & \textbf{255.5} & 67.93 & 51.16 & 62.34 & 44.38 & 57.85 \\
\rowcolor{oursrow}\quad \textbf{\method (Pretrained Agent)} & Qwen3-8B & 408.4 & 72.82 & 56.23 & 67.29 & 40.01 & 60.47 \\
\rowcolor{oursrow}\quad \textbf{\method (\rlmethod)} & Qwen3-8B & 409.2 & \textbf{73.06} & \textbf{66.84} & \textbf{70.99} & \textbf{58.43} & \textbf{67.85} \\
\bottomrule
\end{tabular}%
}
\vspace{2pt}
\par{\scriptsize $^\dagger$9-task cascades evaluated on FAR under the official causal probe scorer. Full 12-subtask results are reported in \cref{tab:ovobench_12tasks}.}

\end{table*}

\begin{table*}[t]
\caption{\textbf{Zero-shot cross-benchmark transfer to StreamingBench and Video-Holmes.} Without target-benchmark fine-tuning, \method improves streaming perception across resolutions and doubles multi-hop reasoning accuracy on Qwen3-VL-8B.}
\label{tab:videoholmes_7tasks}
\centering
\footnotesize
\renewcommand{\arraystretch}{0.96}
\setlength{\tabcolsep}{4.5pt}
\resizebox{\linewidth}{!}{%
\begin{tabular}{@{}l rrr rrr@{}}
\toprule
 & \multicolumn{3}{c}{\textbf{Qwen2.5-VL-7B-Instruct}} & \multicolumn{3}{c}{\textbf{Qwen3-VL-8B-Instruct}} \\
\cmidrule(lr){2-4} \cmidrule(l){5-7}
Benchmark \& Capability Track & SimpleStream & \textbf{\method} & $\Delta$ (pp) & SimpleStream & \textbf{\method} & $\Delta$ (pp) \\
\midrule
\quad StreamingBench: Real-Time Visual (RTVU, 10 tasks) & 56.00 & \cellcolor{oursrow}\textbf{74.00} & \cellcolor{oursrow}\textbf{+18.00} & 74.00 & \cellcolor{oursrow}\textbf{77.50} & \cellcolor{oursrow}\textbf{+3.50} \\
\quad StreamingBench: Omni-Source (OMNI, 6 tasks) & 51.67 & \cellcolor{oursrow}\textbf{67.50} & \cellcolor{oursrow}\textbf{+15.83} & 53.33 & \cellcolor{oursrow}\textbf{65.83} & \cellcolor{oursrow}\textbf{+12.50} \\
\quad StreamingBench: Contextual \& Proactive (CU, 2 tasks) & 22.50 & \cellcolor{oursrow}\textbf{22.50} & \cellcolor{oursrow}0.00 & 12.50 & \cellcolor{oursrow}\textbf{32.50} & \cellcolor{oursrow}\textbf{+20.00} \\
\rowcolor{oursrow}\quad \textbf{StreamingBench Overall ($224^2$, $255.5$ base tok)} & 50.83 & \textbf{66.11} & \textbf{+15.28} & 60.28 & \textbf{68.61} & \textbf{+8.33} \\
\rowcolor{oursrow}\quad \textbf{StreamingBench Overall ($672^2$, $2,299$ base tok)} & 45.28 & \textbf{74.17} & \textbf{+28.89} & 53.89 & \textbf{72.50} & \textbf{+18.61} \\
\midrule
\quad Video-Holmes: UniformHistory-8f ($512$ tok) & 27.86 & \cellcolor{oursrow}\textbf{32.14} & \cellcolor{oursrow}\textbf{+4.28} & 33.57 & \cellcolor{oursrow}\textbf{47.14} & \cellcolor{oursrow}\textbf{+13.57} \\
\rowcolor{oursrow}\quad \textbf{Video-Holmes Overall ($224^2$, 7 tasks)} & 17.14 & \textbf{32.14} & \textbf{+15.00} & 23.57 & \textbf{47.14} & \textbf{+23.57} \\
\bottomrule
\end{tabular}%
}

\end{table*}

\subsection{Tool-Utility Policy Optimization}
\label{sec:method_rl}

\textbf{Tool-Utility Policy Optimization (\rlmethod)} trains selective tool use through counterfactual rewards (\cref{fig:architecture}c). It combines balanced training contexts with a utility signal for each evidence-seeking action.

\paragraph{Training procedure.}
Supervised fine-tuning (SFT) initializes the agent with trajectories containing diagnostic text, tool calls, evidence cards, and final answers. Cross-video negative retrieval pairs supervise \emph{None} cards. This establishes the interaction protocol for routing and synthesis. \rlmethod then optimizes the policy using task correctness, evidence utility, and visual cost. The verifier shares the agent's weights, while a frozen reference model scores the counterfactual gain. This separates learning the output protocol from improving the evidence acquisition policy.

\paragraph{Balanced context perturbation.}
Coarse frames often suffice for guessing on standard training clips. We therefore sample a balanced mix of clean contexts, reduced spatial detail, missing history, and incomplete actions. These settings provide useful training cases for \ActAnswer{}, \ActZoom{}, \ActRecall{}, and \ActWait{}. Each prompt uses the same perturbed context for all $G$ rollouts in its group. The exact mixture and perturbations are specified in \cref{app:implementation_details}.

\paragraph{Counterfactual tool-utility reward.}
We evaluate the correct answer $y^*$ under the frozen reference model $\piref$ before and after acquiring evidence $E_{i,t}$. For waiting, we compare the recent windows at query time $t_q$ and after an interval $\Delta t$. The difference in log-likelihood measures the tool's utility. We normalize this gain by answer length to obtain
\begin{equation}
\label{eq:tool_utility_gain}
\DeltaI_i = \frac{1}{\abs{y^*}} \begin{cases}
\log \frac{\piref(y^* \mid \Wrec, E_{i,t}, q_t)}{\piref(y^* \mid \Wrec, q_t)}, & \text{Zoom/Recall}, \\[4pt]
\log \frac{\piref(y^* \mid \mathcal{W}_{t_q + \Delta t}^{\mathrm{rec}}, q_t)}{\piref(y^* \mid \mathcal{W}_{t_q}^{\mathrm{rec}}, q_t)}, & \text{Wait}, \\[2pt]
0, & \text{Answer}.
\end{cases}
\end{equation}
Answer scoring uses a single teacher-forced prefill over the short target answer. A positive utility reward also requires a correct final answer and diagnostic text that matches the selected action. We check this match using deterministic cue rules (\cref{app:action_consistency}). These conditions gate the clipped gain to produce $\RIG$.

\paragraph{Composite reward and turn-balanced optimization.}
For each rollout $\tau_i$, the reward combines answer accuracy $\Racc$, tool utility $\RIG$, format validity $\Ranchor$, and normalized visual token cost $C_i$. We compute group-centered advantages without variance normalization \citep{liu2025drgrpo}.
\begin{equation}
\label{eq:composite_reward}
\begin{aligned}
R_i &= \Racc + \lambda_{\mathrm{ig}} \RIG + \lambda_{\mathrm{fmt}} \Ranchor - \lambda_{\mathrm{cost}} C_i, \\
\hat{A}_i &= R_i - \frac{1}{G} \sum_{j=1}^G R_j,
\end{aligned}
\end{equation}
We average token losses within each turn and weight routing and synthesis turns equally \citep{zhou2026turnwise}. This prevents a longer answer from dominating the routing update. At inference, the agent commits to an \ActAnswer{} prediction when its confidence is at least $\tau_{\mathrm{conf}}$. A low-confidence direct answer is escalated to the highest-scoring tool (\cref{alg:forager}). Training settings are listed in \cref{tab:supp_hyperparams}.

\section{Experiments}
\label{sec:experiments}

\begin{figure*}[t]
    \centering
    \includegraphics[width=\textwidth]{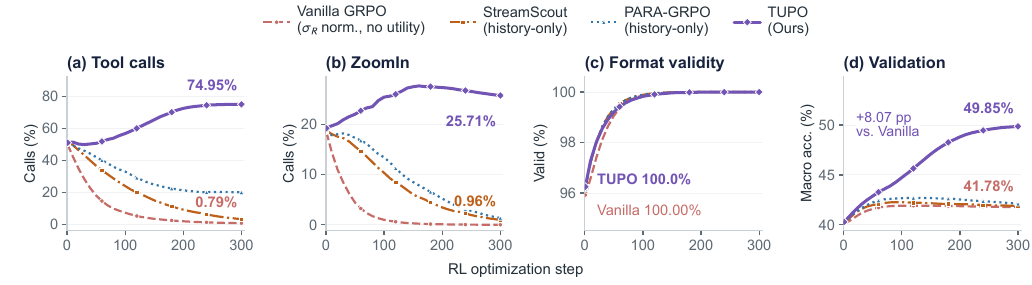}
    \caption{\textbf{Multi-tool RL training dynamics across optimization recipes.} Standard and history-only RL baselines suffer severe tool collapse, whereas \rlmethod sustains all three foraging tools and achieves the highest validation accuracy.}
    \label{fig:rl_dynamics}
\end{figure*}

\begin{table*}[t]
\caption{\textbf{Component ablations of tool routing and evidence fusion on OVO-Bench.} Each action improves its target capability. Evidence gating and SDAM suppress false-premise errors while keeping the visual context compact.}
\label{tab:ablation_tools}
\label{tab:ablation_fusion}
\centering
\footnotesize
\renewcommand{\arraystretch}{0.95}
\setlength{\tabcolsep}{4.0pt}
\resizebox{\linewidth}{!}{%
\begin{tabular}{@{}l r rrrrr r@{}}
\toprule
 & & \multicolumn{5}{c}{\textbf{OVO-Bench Accuracy (\%)}} & \\
\cmidrule(lr){3-7}
Ablation Variant & \textbf{Vis.\ Tokens} & \textbf{RT} & \textbf{BT} & \textbf{HLD} & \textbf{FAR} & \textbf{Overall} & \textbf{$\Delta$ Overall} \\
\midrule
\multicolumn{8}{@{}l}{\emph{Ablation A: Four-Way Tool Routing (Same Trained Qwen2.5-VL-7B Checkpoint)}} \\
\quad 1. Answer Only (Same-Ckpt SimpleStream-4f Origin) & 255.5 & 55.60 & 57.86 & 86.56 & 36.60 & 51.41 & 0.00 \\
\quad 2. + \ActZoom{} Only (Present Spatial/FPS) & 286.9 & \textbf{66.27} & 57.86 & 86.56 & 36.60 & 56.75 & +5.34 \\
\quad 3. + \ActRecall{} Only (Past Event Archive) & 263.7 & 55.60 & \textbf{67.11} & \textbf{99.46} & 40.23 & 54.63 & +3.22 \\
\quad 4. + \ActWait{} Only (Future Stream Deferral) & 255.5 & 55.60 & 57.86 & 86.56 & 47.02 & 54.02 & +2.61 \\
\rowcolor{oursrow}\quad 5. \textbf{Full \method (Ours)} & \textbf{295.1} & \textbf{66.27} & \textbf{67.11} & \textbf{99.46} & \textbf{50.65} & \textbf{62.57} & \textbf{+11.16} \\
\midrule
\multicolumn{8}{@{}l}{\emph{Ablation B: Evidence Gating and SDAM (Qwen2.5-VL-7B)}} \\
\quad 1. SimpleStream-4f (No Historical Context) & 255.5 & 55.60 & 57.86 & 86.56 & 36.60 & 51.41 & 0.00 \\
\quad 2. Naive 8-Frame Raw Concat (OASIS-style) & 765.3 & 53.94 & 47.57 & 29.03 & 41.10 & 49.14 & -2.27 \\
\quad 3. Sub-Agent \EvCard{} (w/o Evidence Gate, w/o SDAM) & 323.8 & 64.33 & 55.96 & 73.12 & \textbf{50.65} & 58.82 & +7.41 \\
\quad 4. + Evidence Gate (w/o SDAM) & 295.1 & 64.33 & 62.77 & 93.55 & \textbf{50.65} & 60.52 & +9.11 \\
\quad 5. + SDAM (w/o Evidence Gate) & 323.8 & \textbf{66.27} & 59.40 & 76.34 & \textbf{50.65} & 60.65 & +9.24 \\
\rowcolor{oursrow}\quad 6. \textbf{\EvCard{} + Evidence Gate + SDAM (Ours)} & \textbf{295.1} & \textbf{66.27} & \textbf{67.11} & \textbf{99.46} & \textbf{50.65} & \textbf{62.57} & \textbf{+11.16} \\
\bottomrule
\end{tabular}%
}

\end{table*}

\begin{figure*}[t]
    \centering
    \includegraphics[width=\textwidth]{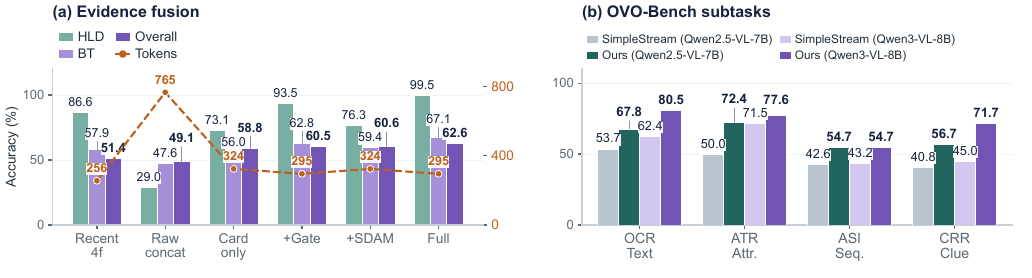}
    \caption{\textbf{Evidence fusion and subtask gains.} Recent 4f denotes SimpleStream-4f. Gate and SDAM each augment Card only.}
    \label{fig:ablation_and_generalization}
\end{figure*}

\subsection{Experimental Setup}
\label{sec:exp_setup}

\paragraph{Benchmarks and protocol.}
We evaluate \method with Qwen2.5-VL-7B \citep{bai2025qwen25vl} and Qwen3-VL-8B \citep{yang2025qwen3vl} on real video streams.
\textbf{OVO-Bench} \citep{niu2025ovobench} evaluates Real-Time Visual Perception (RT), Backward Tracing (BT), and Forward Active Responding (FAR). FAR responses are scored at causal probe timestamps (\cref{app:protocols,app:wait_policy}).
\textbf{StreamingBench} \citep{lin2024streamingbench} tests real-time, omni-source, and contextual understanding. \textbf{Video-Holmes} \citep{cheng2025videoholmes} evaluates multi-hop film reasoning (\cref{app:subtasks}).

\emph{Pretrained Agent} denotes tool use with the original VLM weights before agent fine-tuning. Transfer results use the \rlmethod-trained agent without fine-tuning on the target benchmarks. We refer to this evaluation as \emph{zero-shot transfer}.

\paragraph{Training data and visual token cost.}
Training clips come from public video corpora including LLaVA-Video-178K, ShareGPT4Video, ActivityNet, and COIN. We deduplicate them against the evaluation benchmarks by video ID and perceptual hash. SFT and reinforcement learning use disjoint prompt sets. A held-out validation split selects the confidence threshold $\tau_{\mathrm{conf}}$. Main-context visual tokens count the recent window and evidence passed to the primary agent, $L_{\mathrm{rec}}+L_{\mathrm{forage}}$. Sub-agent verification cost, CPU memory, and latency are reported separately in \cref{tab:supp_routing_ledger}.

\subsection{Main Results on OVO-Bench}
\label{sec:exp_ovobench}

\finding{On-demand foraging improves perception and memory together.}{\method improves current-scene perception, historical recall, and proactive response with a compact visual context.}

\paragraph{Comparison against memory banks and history-retrieval cascades.}
\Cref{tab:ovobench_tracks} compares \method with memory banks and history-retrieval cascades across both resolution settings. Memory banks retain historical context continuously, while cascades retrieve it after an uncertain initial answer. \method keeps the initial window compact and selects additional evidence for each query. Compared with SimpleStream-4f, it improves perception, recall, and proactive response on both backbones. At high resolution, Qwen3-VL-8B reaches \textbf{71.30\%} overall accuracy while using 64\% fewer main-context visual tokens than StreamScout-R.

\paragraph{Tool use before and after training.}
Training activates spatial inspection on Qwen2.5-VL-7B and improves overall accuracy on both backbones (\cref{tab:supp_routing_ledger,tab:ovobench_tracks}). Qwen3-VL-8B achieves its gain while maintaining a similar overall \ActZoom{} invocation rate.

\paragraph{Targeted gains across spatial, historical, and proactive subtasks.}
Fine-grained results show how each action addresses a different evidence need (\cref{fig:ablation_and_generalization}b, \cref{tab:ovobench_12tasks}). \ActZoom{} improves scene text and fine-attribute recognition by recovering spatial detail. Verification and evidence gating support historical recall and reject false-premise events. Combined with SDAM, the full agent reaches 99.46\% accuracy on HLD. On proactive tasks, \ActWait{} provides further observations before commitment. Learned routing outperforms fixed delays and readiness thresholds (\cref{tab:supp_progression_controls}).

\paragraph{Routing across input resolutions.}
For both backbones, the high-resolution setting has a larger direct-answer share and a smaller \ActZoom{} share (\cref{tab:supp_routing_ledger}). Detailed current-scene observations are already available in the base window. Recall and waiting remain active in this setting. They target earlier events or later observations that additional spatial detail cannot supply. The tool mix thus varies with the evidence available at query time.

\paragraph{Where additional visual tokens are spent.}
\Cref{tab:supp_routing_ledger} separates the cost of the recent window from that of invoked tools. The recent window contributes to every query. At low resolution, \ActZoom{} contributes more added main-context tokens than \ActRecall{} on both backbones. Verified recall keeps its return compact through a single anchor frame. Candidate inspection takes place in the sub-agent context, whose visual tokens are reported separately. For a \emph{Found} result, the main agent synthesizes its answer using the factual note and verified anchor.

\subsection{Generalization to StreamingBench and Video-Holmes}
\label{sec:exp_videoholmes}

\finding{Selective foraging transfers zero-shot across benchmarks.}{\method generalizes across resolutions on StreamingBench and doubles Qwen3-VL-8B multi-hop reasoning accuracy on Video-Holmes.}

\begin{figure}[t]
    \begin{minipage}[c]{0.52\linewidth}
        \includegraphics[width=\linewidth]{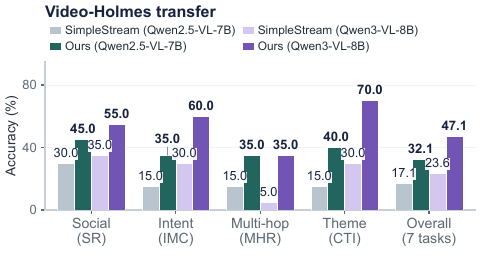}
    \end{minipage}\hfill
    \begin{minipage}[c]{0.43\linewidth}
        \captionsetup{skip=0pt}
        \caption{\textbf{Zero-shot transfer to Video-Holmes.} Ours improves all displayed categories. Overall covers all seven tasks.}
        \label{fig:videoholmes_transfer}
    \end{minipage}
\end{figure}

\paragraph{Cross-benchmark streaming and multi-hop transfer.}
\Cref{tab:videoholmes_7tasks} evaluates zero-shot transfer to StreamingBench and Video-Holmes. On StreamingBench, the gains hold across both resolution settings. Video-Holmes requires connecting evidence across film scenes. \method retrieves relevant historical segments, verifies them through the sub-agent, and uses SDAM during synthesis. This doubles Qwen3-VL-8B accuracy relative to the short-window baseline (\cref{fig:videoholmes_transfer}).

\paragraph{Adaptive routing across diverse video horizons.}
Routing adapts to the evidence needed by each query. Clear foreground actions often support a direct \ActAnswer{}. Dense scene text and small objects trigger \ActZoom{}. Questions about motives and plot favor \ActRecall{} to connect earlier events. The same policy is used across benchmarks without target-specific tuning.

\begin{figure*}[t]
    \centering
    \includegraphics[width=\textwidth]{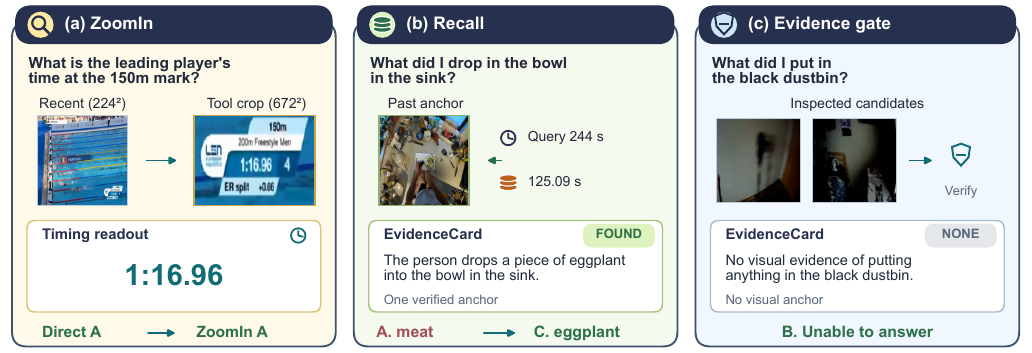}
    \caption{\textbf{Qualitative analysis.} Controlled tool probes inspect fine detail and verify past events.}
    \label{fig:evidence_cases}
\end{figure*}

\subsection{Ablation Studies and Mechanism Analysis}
\label{sec:exp_ablations}

\finding{Evidence selection and tool training each improve accuracy.}{Evidence gating and SDAM reduce false-premise hallucinations, while \rlmethod sustains spatial inspection.}

\paragraph{Tool contributions and evidence fusion.}
\Cref{tab:ablation_tools} (Ablation~A) enables each tool separately on the same trained checkpoint. A disabled tool falls back to direct \ActAnswer{}. Each tool strengthens its corresponding capability, and the full set gives the best overall result in this ablation. Ablation~B varies evidence gating and SDAM while keeping model weights and initial routes fixed (\cref{tab:ablation_fusion}). Raw historical concatenation expands the context and can induce false-premise errors. Evidence gating removes unverified frames, while SDAM controls attention when additional visual evidence is present. Their combination gives the strongest HLD result (\cref{fig:ablation_and_generalization}a).

\paragraph{Multi-tool RL training dynamics.}
\Cref{fig:rl_dynamics} compares RL recipes from the same initial checkpoint. Standard cost-penalized GRPO suppresses tool use when direct guessing receives higher group-centered advantage (\cref{app:proof_crowding}). History-only RL keeps retrieval active but can leave spatial inspection without a useful reward signal \citep{zhang2026streamscout,yang2026paravt}. \rlmethod measures each tool's contribution through the counterfactual gain $\DeltaI$. Balanced contexts expose cases that need different tools, and turn-balanced updates preserve the routing signal. The resulting policy sustains all three tools.

\paragraph{Effect of tool-utility rewards.}
\Cref{tab:utility_reward} compares \rlmethod with matched training that removes $\RIG$. Both settings retain the same SFT initialization, context perturbations, turn weighting, and confidence calibration. Without the utility term, spatial inspection disappears on Qwen2.5-VL-7B. Adding it retains \ActZoom{} and improves proactive and overall accuracy on both backbones. The log-likelihood gain provides feedback on the evidence returned by each tool (\cref{eq:tool_utility_gain}). This links tool selection to observations that support the answer while retaining the visual cost penalty.

\begin{table}[t]
\caption{\textbf{Effect of the tool-utility reward on OVO-Bench.} SFT initialization, context perturbations, turn weighting, and calibration are held fixed. ZoomIn reports invocation rate. FAR and Overall report accuracy.}
\label{tab:utility_reward}
\centering
\footnotesize
\renewcommand{\arraystretch}{1.05}
\setlength{\tabcolsep}{3.5pt}
\begin{tabular}{@{}l rrr@{}}
\toprule
Training Objective & \ActZoom{} (\%) & FAR & Overall \\
\midrule
\multicolumn{4}{@{}l}{\emph{Qwen2.5-VL-7B}} \\
w/o utility reward & 0.00 & 34.67 & 57.83 \\
\rowcolor{oursrow}\textbf{\rlmethod} & \textbf{12.26} & \textbf{50.65} & \textbf{62.57} \\
\addlinespace[2pt]
\multicolumn{4}{@{}l}{\emph{Qwen3-VL-8B}} \\
w/o utility reward & 15.06 & 42.48 & 62.31 \\
\rowcolor{oursrow}\textbf{\rlmethod} & \textbf{25.79} & \textbf{58.43} & \textbf{67.85} \\
\bottomrule
\end{tabular}

\end{table}

\paragraph{Selective waiting versus fixed delays.}
\Cref{tab:supp_progression_controls} compares learned waiting with fixed delays and a readiness threshold applied without learned routing. A fixed delay postpones every answer, including queries already supported by the current frames. It can also remain too short for a longer action. Applying a readiness threshold to every query introduces unnecessary waits. Learned \ActWait{} routing selects the queries that need more observation.

\subsection{Qualitative Analysis}
\label{sec:exp_cases}

\Cref{fig:evidence_cases} compares direct answers with controlled tool probes on OVO-Bench using Qwen2.5-VL-7B.

\ActZoom{} makes the swimming timer readable in the returned view. For the earlier kitchen event, a verified anchor changes the answer from \emph{meat} to \emph{eggplant}. For the dustbin query, the verifier finds no support in the inspected candidates and returns \emph{None}. The main agent receives no historical visual anchor and answers \emph{Unable to answer}.

\FloatBarrier
\section{Conclusion}
\label{sec:conclusion}

\method combines present perception, historical recall, and proactive response through on-demand evidence acquisition. It keeps a compact window of recent frames and gathers missing evidence through \ActZoom{}, \ActRecall{}, or \ActWait{}. The same VLM verifies retrieved history in a separate context. SDAM isolates visual streams during synthesis, and \rlmethod learns selective tool use through counterfactual utility rewards. On OVO-Bench, \method reaches 71.3\% overall accuracy on Qwen3-VL-8B with 64\% fewer main-context visual tokens than StreamScout-R. Zero-shot transfer to StreamingBench and Video-Holmes supports this design across streaming perception and multi-hop reasoning.

\section*{LLM Usage Statement}
We used large language model assistants to polish author-written manuscript text and refactor evaluation scripts. For the manuscript, assistance was restricted to grammar refinement, sentence simplification, and table formatting. These tools were not used for literature selection, research ideation, experimental design, data generation, or scientific conclusions. All experiments were executed on real video streams using the evaluated vision-language models. The authors take full responsibility for the manuscript, code, and empirical findings.

\clearpage
\pdfbookmark[1]{References}{references}
{%
  \setlength{\bibsep}{1pt plus 0.5pt minus 0.5pt}
  \renewcommand{\bibfont}{\fontsize{9}{10.5}\selectfont}
  \bibliographystyle{ieeenat_fullname}
  \bibliography{references}
}

\clearpage
\pdfbookmark[0]{Appendix}{appendix}
\beginappendix
\crefalias{section}{appendix}
\suppressfloats[t]
\section{Complete Subtask Breakdowns}
\label{app:subtasks}

\Cref{tab:ovobench_12tasks} reports the complete 12-subtask results on OVO-Bench corresponding row for row to \cref{tab:ovobench_tracks} in the main paper. Across both resolution regimes and both VLM backbones, \ActZoom{} drives gains on spatial text and attribute subtasks (OCR, ATR, FPD), sub-agent \ActRecall{} with SDAM suppresses false-premise errors on Hallucination Detection (HLD) while improving episodic recall (EPM, ASI), and \ActWait{} lifts proactive streaming accuracy across repetition counting, state transitions, and clue reveal (REC, SSR, CRR). \Cref{tab:supp_videoholmes} provides the per-subtask transfer breakdown on Video-Holmes alongside working-memory, episodic-memory, and proactive slices on StreamingBench, confirming that selective episode verification consistently outperforms uniform multi-frame history baselines.

\begin{table*}[t]
\caption{\textbf{Complete 12-subtask accuracy breakdown on OVO-Bench.} On-demand foraging yields consistent gains across spatial perception, historical recall, and proactive response tracks, matching \cref{tab:ovobench_tracks} row for row.}
\label{tab:ovobench_12tasks}
\centering
\footnotesize
\renewcommand{\arraystretch}{1.05}
\setlength{\tabcolsep}{3.2pt}
\resizebox{\textwidth}{!}{%
\begin{tabular}{@{}ll rrrrrr r rrr r r rrr r r@{}}
\toprule
 & & \multicolumn{7}{c}{\textbf{Real-Time Visual Perception (6 Subtasks)}} & \multicolumn{4}{c}{\textbf{Backward Tracing (3)}} & & \multicolumn{4}{c}{\textbf{Forward Active (3)}} & \\
\cmidrule(lr){3-9} \cmidrule(lr){10-13} \cmidrule(lr){15-18}
Method & Backbone & OCR & ACR & ATR & STU & OJR & FPD & \textbf{RT} & EPM & ASI & HLD & \textbf{BT} & \textbf{9-Avg} & REC & SSR & CRR & \textbf{FAR} & \textbf{Overall} \\
\midrule
\multicolumn{19}{@{}l}{\emph{Prior Streaming Memory Systems}} \\
\quad Flash-VStream \citep{zhang2024flashvstream} & 7B & 24.16 & 29.36 & 28.45 & 33.71 & 28.80 & 25.74 & 28.37 & 39.06 & 37.16 & 5.91 & 27.38 & 28.04 & 8.02 & 67.25 & 60.00 & 45.09 & 32.30 \\
\quad Dispider \citep{qian2025dispider} & 7B & 57.72 & 49.54 & 62.07 & 44.94 & 51.63 & 61.39 & 54.55 & 48.48 & 55.41 & 4.30 & 36.06 & 48.39 & 18.05 & 37.36 & 48.75 & 34.72 & 44.97 \\
\quad StreamForest \citep{zeng2025streamforest} & Qwen2.5-7B & 68.46 & 53.21 & 71.55 & 47.75 & 60.87 & 65.35 & 61.20 & 58.92 & 64.86 & 32.26 & 52.01 & 58.14 & 32.81 & \textbf{70.59} & \textbf{57.08} & \textbf{53.49} & 56.98 \\
\quad HERMES$^\dagger$ \citep{li2025hermes} & Qwen2.5-7B & \textbf{85.23} & \textbf{64.22} & \textbf{71.55} & \textbf{53.37} & \textbf{65.22} & \textbf{74.26} & \textbf{68.98} & 48.48 & 62.16 & \textbf{37.63} & 49.42 & \textbf{62.46} & 24.64 & 58.66 & 40.00 & 41.10 & \textbf{57.12} \\
\midrule
\multicolumn{19}{@{}l}{\emph{High-Resolution Streaming Regime ($672^2$ Base Window)}} \\
\quad OASIS$^\dagger$ \citep{liang2026oasis} & Qwen2.5-7B & 85.23 & 72.48 & 66.38 & 52.25 & 64.67 & 67.33 & 68.06 & 51.85 & 58.78 & 48.92 & 53.18 & 63.10 & 24.07 & 57.71 & 39.58 & 40.45 & 57.44 \\
\quad StreamScout-R$^\dagger$ \citep{zhang2026streamscout} & Qwen2.5-7B & \textbf{92.62} & 77.98 & \textbf{77.59} & 57.30 & \textbf{78.26} & 75.25 & 76.50 & \textbf{55.22} & \textbf{64.19} & 89.25 & 69.55 & 74.18 & 21.78 & 61.05 & 37.92 & 40.25 & 65.70 \\
\quad SimpleStream-4f ($672^2$) \citep{shen2026simplestream} & Qwen2.5-7B & 65.77 & 68.81 & 57.76 & 47.75 & 67.39 & 69.31 & 62.80 & 49.16 & 40.54 & 52.69 & 47.46 & 57.69 & 17.62 & 51.35 & 40.83 & 36.60 & 52.41 \\
\rowcolor{oursrow}\quad \textbf{\method ($672^2$ Native Hi-Res)} & Qwen2.5-7B & 89.26 & \textbf{78.90} & \textbf{77.59} & \textbf{61.80} & 77.72 & \textbf{77.23} & \textbf{77.08} & 54.88 & 56.08 & \textbf{99.46} & \textbf{70.14} & \textbf{74.77} & \textbf{28.51} & \textbf{66.77} & \textbf{56.67} & \textbf{50.65} & \textbf{68.74} \\
\addlinespace[1.5pt]
\quad OASIS$^\dagger$ \citep{liang2026oasis} & Qwen3-8B & 92.62 & 81.65 & 78.45 & 65.17 & 78.26 & 68.00 & 77.36 & 58.25 & 67.57 & 51.61 & 59.14 & 71.29 & 26.93 & 69.79 & 46.67 & 47.80 & 65.41 \\
\quad StreamScout-R$^\dagger$ \citep{zhang2026streamscout} & Qwen3-8B & \textbf{93.96} & 84.40 & \textbf{80.17} & 67.42 & \textbf{78.80} & \textbf{74.26} & \textbf{79.84} & \textbf{58.59} & \textbf{69.59} & \textbf{98.39} & \textbf{75.52} & \textbf{78.40} & 27.22 & 65.66 & 45.00 & 45.96 & 70.29 \\
\quad SimpleStream-4f ($672^2$) \citep{shen2026simplestream} & Qwen3-8B & 87.25 & 78.90 & 75.00 & 65.17 & 74.46 & 73.27 & 75.67 & 50.84 & 47.30 & 58.60 & 52.25 & 67.86 & 16.91 & 71.22 & 45.00 & 44.38 & 61.99 \\
\rowcolor{oursrow}\quad \textbf{\method ($672^2$ Native Hi-Res)} & Qwen3-8B & 89.93 & \textbf{85.32} & 78.45 & \textbf{69.10} & 76.09 & 69.31 & 78.03 & 54.21 & 62.16 & 95.70 & 70.69 & 75.58 & \textbf{32.23} & \textbf{71.38} & \textbf{71.67} & \textbf{58.43} & \textbf{71.30} \\
\midrule
\multicolumn{19}{@{}l}{\emph{Budget-Constrained Streaming Regime ($224^2$ Base Window)}} \\
\quad SimpleStream-4f ($224^2$) \citep{shen2026simplestream} & Qwen2.5-7B & 53.69 & 66.06 & 50.00 & 43.82 & 63.59 & 56.44 & 55.60 & 44.44 & 42.57 & 86.56 & 57.86 & 56.35 & 17.62 & 51.35 & 40.83 & 36.60 & 51.41 \\
\rowcolor{oursrow}\quad \textbf{\method (Pretrained Agent)} & Qwen2.5-7B & 61.07 & \textbf{71.56} & 65.52 & 47.75 & \textbf{66.85} & 69.31 & 63.68 & \textbf{49.49} & \textbf{55.41} & 72.04 & 58.98 & 62.11 & 18.48 & 47.38 & 40.00 & 35.29 & 55.41 \\
\rowcolor{oursrow}\quad \textbf{\method (\rlmethod)} & Qwen2.5-7B & \textbf{67.79} & 68.81 & \textbf{72.41} & \textbf{50.00} & 66.30 & \textbf{72.28} & \textbf{66.27} & 47.14 & 54.73 & \textbf{99.46} & \textbf{67.11} & \textbf{66.55} & \textbf{28.51} & \textbf{66.77} & \textbf{56.67} & \textbf{50.65} & \textbf{62.57} \\
\addlinespace[1.5pt]
\quad SimpleStream-4f ($224^2$) \citep{shen2026simplestream} & Qwen3-8B & 62.42 & \textbf{76.15} & 71.55 & 60.11 & 69.02 & 68.32 & 67.93 & 46.80 & 43.24 & 63.44 & 51.16 & 62.34 & 16.91 & 71.22 & 45.00 & 44.38 & 57.85 \\
\rowcolor{oursrow}\quad \textbf{\method (Pretrained Agent)} & Qwen3-8B & \textbf{80.54} & 74.31 & 76.72 & 61.24 & \textbf{72.83} & \textbf{71.29} & 72.82 & 48.48 & \textbf{56.76} & 63.44 & 56.23 & 67.29 & 16.05 & 61.05 & 42.92 & 40.01 & 60.47 \\
\rowcolor{oursrow}\quad \textbf{\method (\rlmethod)} & Qwen3-8B & \textbf{80.54} & 74.31 & \textbf{77.59} & \textbf{61.80} & \textbf{72.83} & \textbf{71.29} & \textbf{73.06} & \textbf{51.18} & 54.73 & \textbf{94.62} & \textbf{66.84} & \textbf{70.99} & \textbf{32.23} & \textbf{71.38} & \textbf{71.67} & \textbf{58.43} & \textbf{67.85} \\
\bottomrule
\end{tabular}%
}
\vspace{2pt}
\par{\scriptsize $^\dagger$9-task cascades evaluated on FAR under the official causal probe scorer ($N=1,640$ total OVO-Bench queries).}

\end{table*}

\begin{table*}[t]
\caption{\textbf{Per-subtask transfer results on Video-Holmes and memory slices on StreamingBench.} Verified episode recall substantially outperforms uniform historical frame sampling across multi-hop reasoning categories.}
\label{tab:supp_videoholmes}
\centering
\footnotesize
\renewcommand{\arraystretch}{1.02}
\setlength{\tabcolsep}{4.3pt}
\resizebox{\linewidth}{!}{%
\begin{tabular}{@{}l rrr rrr@{}}
\toprule
 & \multicolumn{3}{c}{\textbf{Qwen2.5-VL-7B-Instruct}} & \multicolumn{3}{c}{\textbf{Qwen3-VL-8B-Instruct}} \\
\cmidrule(lr){2-4} \cmidrule(l){5-7}
Benchmark / Subtask & SimpleStream & \textbf{\method} & $\Delta$ (pp) & SimpleStream & \textbf{\method} & $\Delta$ (pp) \\
\midrule
\multicolumn{7}{@{}l}{\emph{Video-Holmes Per-Subtask Breakdown (7 Multi-Hop Reasoning Subtasks)}} \\
\quad Social Reasoning (SR) & 30.00 & \cellcolor{oursrow}\textbf{45.00} & \cellcolor{oursrow}\textbf{+15.00} & 35.00 & \cellcolor{oursrow}\textbf{55.00} & \cellcolor{oursrow}\textbf{+20.00} \\
\quad Intention \& Motive (IMC) & 15.00 & \cellcolor{oursrow}\textbf{35.00} & \cellcolor{oursrow}\textbf{+20.00} & 30.00 & \cellcolor{oursrow}\textbf{60.00} & \cellcolor{oursrow}\textbf{+30.00} \\
\quad Temporal Causal (TCI) & 25.00 & \cellcolor{oursrow}\textbf{30.00} & \cellcolor{oursrow}\textbf{+5.00} & 30.00 & \cellcolor{oursrow}\textbf{40.00} & \cellcolor{oursrow}\textbf{+10.00} \\
\quad Temporal Association (TA) & 10.00 & \cellcolor{oursrow}\textbf{15.00} & \cellcolor{oursrow}\textbf{+5.00} & 5.00 & \cellcolor{oursrow}\textbf{30.00} & \cellcolor{oursrow}\textbf{+25.00} \\
\quad Multi-Hop Reasoning (MHR) & 15.00 & \cellcolor{oursrow}\textbf{35.00} & \cellcolor{oursrow}\textbf{+20.00} & 5.00 & \cellcolor{oursrow}\textbf{35.00} & \cellcolor{oursrow}\textbf{+30.00} \\
\quad Plot Analysis (PAR) & 10.00 & \cellcolor{oursrow}\textbf{25.00} & \cellcolor{oursrow}\textbf{+15.00} & 30.00 & \cellcolor{oursrow}\textbf{40.00} & \cellcolor{oursrow}\textbf{+10.00} \\
\quad Character \& Theme (CTI) & 15.00 & \cellcolor{oursrow}\textbf{40.00} & \cellcolor{oursrow}\textbf{+25.00} & 30.00 & \cellcolor{oursrow}\textbf{70.00} & \cellcolor{oursrow}\textbf{+40.00} \\
\quad UniformHistory-12f ($768$ tok) & 22.86 & \cellcolor{oursrow}\textbf{32.14} & \cellcolor{oursrow}\textbf{+9.28} & 10.71 & \cellcolor{oursrow}\textbf{47.14} & \cellcolor{oursrow}\textbf{+36.43} \\
\quad UniformHistory-8f ($512$ tok) & 27.86 & \cellcolor{oursrow}\textbf{32.14} & \cellcolor{oursrow}\textbf{+4.28} & 33.57 & \cellcolor{oursrow}\textbf{47.14} & \cellcolor{oursrow}\textbf{+13.57} \\
\rowcolor{oursrow}\quad \textbf{Video-Holmes Overall (7 Subtasks)} & 17.14 & \textbf{32.14} & \textbf{+15.00} & 23.57 & \textbf{47.14} & \textbf{+23.57} \\
\midrule
\multicolumn{7}{@{}l}{\emph{StreamingBench Memory \& Proactive Slices ($224^2$ Budgeted Window)}} \\
\quad Working Memory (WM, Immediate Perception) & 53.33 & \cellcolor{oursrow}\textbf{74.17} & \cellcolor{oursrow}\textbf{+20.84} & 76.67 & \cellcolor{oursrow}\textbf{77.50} & \cellcolor{oursrow}\textbf{+0.83} \\
\quad Episodic Memory (EM, Historical Recall) & 54.09 & \cellcolor{oursrow}\textbf{67.27} & \cellcolor{oursrow}\textbf{+13.18} & 56.82 & \cellcolor{oursrow}\textbf{65.91} & \cellcolor{oursrow}\textbf{+9.09} \\
\quad Proactive Output (PO, Blind Deferral) & 0.00 & \cellcolor{oursrow}\textbf{5.00} & \cellcolor{oursrow}\textbf{+5.00} & 0.00 & \cellcolor{oursrow}\textbf{45.00} & \cellcolor{oursrow}\textbf{+45.00} \\
\bottomrule
\end{tabular}%
}

\end{table*}

\section{Inference and Evaluation Details}
\label{app:algorithm}
\label{app:protocols}
\label{app:controls}

\subsection{Turn-Balanced Objective and Streaming Inference}
\rlmethod uses the group-centered advantages $\hat A_i$ from \cref{eq:composite_reward}. We average token losses within each turn and weight the $K_i\in\{1,2\}$ turns equally \citep{liu2025drgrpo,zhou2026turnwise}. The objective is
\begin{equation}
\label{eq:turn_mean_loss}
\begin{aligned}
\mathcal{J}(\theta) = \E \Bigg[ \frac{1}{G} \sum_{i=1}^G \frac{1}{K_i} \sum_{k=1}^{K_i} & \frac{1}{\abs{y_{i,k}}} \sum_{u=1}^{\abs{y_{i,k}}} \Big( \tilde{\mathcal{L}}_{i,k,u}^{\mathrm{PPO}}(\theta) \\[-2pt]
& - \beta_{\mathrm{KL}} \KL(\pitheta \parallel \piref) \Big) \Bigg].
\end{aligned}
\end{equation}
\Cref{alg:forager} summarizes routing and synthesis at each query. Waiting advances the observed prefix before routing is evaluated again.

\begin{algorithm}[t]
\caption{\method Causal Streaming Inference}
\label{alg:forager}
\renewcommand{\arraystretch}{1.12}
\begin{algorithmic}[1]
\REQUIRE Observed prefix $\Vpref$, query $q_t$, live stream access, policy $\pitheta$, confidence threshold $\tau_{\mathrm{conf}}$.
\STATE Initialize waiting count $n_{\mathrm{wait}} \gets 0$.
\WHILE{true}
    \STATE Update the recent window, CPU ring buffer, and history archive (\cref{app:implementation_details}).
    \STATE Decode action $a_1$ from $(\Wrec,q_t)$ and compute direct-answer confidence $p_{\max}$.
    \IF{$a_1 = \ActAnswer(\hat{y})$ \AND $p_{\max} \ge \tau_{\mathrm{conf}}$}
        \RETURN $\hat{y}$ immediately.
    \ELSIF{$a_1 = \ActAnswer(\hat{y})$ \AND $p_{\max} < \tau_{\mathrm{conf}}$}
        \STATE Set $a_1$ to the highest-logit tool in $\{\ActZoom,\ActRecall,\ActWait\}$.
    \ENDIF
    \IF{$a_1 = \ActWait$}
        \IF{stream access is unavailable or $n_{\mathrm{wait}} \ge 5$}
            \STATE Force a direct answer $\hat{y}$ from the current window.
            \RETURN $\hat{y}$.
        \ENDIF
        \STATE Advance by the requested interval and ingest new frames.
        \STATE $n_{\mathrm{wait}} \gets n_{\mathrm{wait}}+1$.
        \IF{no new frames arrive or $n_{\mathrm{wait}} = 5$}
            \STATE Force a direct answer $\hat{y}$ from the updated window.
            \RETURN $\hat{y}$.
        \ENDIF
    \ELSE
        \IF{$a_1 = \ActZoom$}
            \STATE Extract a high-resolution crop or dense frame burst as $E_t$.
        \ELSIF{$a_1 = \ActRecall$}
            \STATE Retrieve temporally separated segments and verify representative frames to obtain $E_t^{\mathrm{card}}$.
        \ENDIF
        \STATE Form the synthesis context $O_t^{(2)}$ with the returned evidence.
        \STATE Construct $\MSDAM$ via \cref{eq:sdam_mask}.
        \STATE Decode $\hat{y} \sim \pitheta(\cdot\mid O_t^{(2)},\MSDAM)$.
        \RETURN $\hat{y}$.
    \ENDIF
\ENDWHILE
\end{algorithmic}
\end{algorithm}

\begin{table*}[t]
\caption{\textbf{Action routing distribution and visual token accounting on OVO-Bench.} Selective invocation keeps both main-context and amortized sub-agent token budgets compact across backbones and resolutions.}
\label{tab:supp_routing_ledger}
\centering
\footnotesize
\renewcommand{\arraystretch}{1.05}
\setlength{\tabcolsep}{3.0pt}
\resizebox{\linewidth}{!}{%
\begin{tabular}{@{}l rrrr rrrr r rr@{}}
\toprule
 & \multicolumn{4}{c}{\textbf{Action Routing (\%)}} & \multicolumn{4}{c}{\textbf{Main-Context Visual Tokens}} & & \multicolumn{2}{c}{\textbf{System Profile}} \\
\cmidrule(lr){2-5} \cmidrule(lr){6-9} \cmidrule(l){11-12}
Backbone \& Regime & \ActAnswer & \ActZoom & \ActRecall & \ActWait & Base & +Zoom & +Recall & \textbf{Total} & \textbf{+Sub} & VRAM & Latency \\
\midrule
\quad Qwen2.5-VL-7B ($224^2$, Pretrained Agent) & 65.55 & 0.06 & 28.42 & 5.97 & 255.4 & 0.2 & 2.9 & 258.6 & +72.6 & 16.4\,GB & 1,157\,ms \\
\rowcolor{oursrow}\quad \textbf{Qwen2.5-VL-7B ($224^2$, \rlmethod)} & 48.78 & 12.26 & 24.02 & 14.94 & 255.5 & 31.4 & 8.2 & \textbf{295.1} & +61.4 & 16.4\,GB & 357\,ms \\
\quad Qwen3-VL-8B ($224^2$, Pretrained Agent) & 44.45 & 25.61 & 18.90 & 11.04 & 255.5 & 147.5 & 5.4 & 408.4 & +48.3 & 18.2\,GB & 1,264\,ms \\
\rowcolor{oursrow}\quad \textbf{Qwen3-VL-8B ($224^2$, \rlmethod)} & 44.45 & 25.79 & 18.66 & 11.10 & 255.5 & 148.6 & 5.1 & \textbf{409.2} & +47.7 & 18.2\,GB & 1,152\,ms \\
\midrule
\rowcolor{oursrow}\quad \textbf{Qwen2.5-VL-7B ($672^2$ Native Hi-Res)} & 79.21 & 0.43 & 9.88 & 10.48 & 2,299.4 & 4.4 & 9.8 & \textbf{2,313.6} & +25.2 & 17.8\,GB & 902\,ms \\
\rowcolor{oursrow}\quad \textbf{Qwen3-VL-8B ($672^2$ Native Hi-Res)} & 73.11 & 4.08 & 11.40 & 11.40 & 2,299.4 & 41.8 & 12.7 & \textbf{2,353.9} & +29.1 & 19.6\,GB & 1,187\,ms \\
\bottomrule
\end{tabular}%
}
\vspace{2pt}
\par{\scriptsize Note: At $224^2$, each \ActZoom{} crop encodes $256$ tokens on 7B ($448^2$) or $576$ tokens on 8B ($672^2$), and each verified \ActRecall{} anchor adds $64$ tokens ($0$ when unverified). At $672^2$, crops encode $1,024$ tokens ($896^2$) and anchors encode $256$ tokens. \emph{+Sub} reports amortized sub-agent verification tokens across all $1,640$ queries.}

\end{table*}

\subsection{Buffer and Training Details}
\label{app:implementation_details}

\paragraph{Buffers and history segments.}
The recent window contains $w=4$ frames at $224^2$, sampled at $1$\,fps. The VLM uses $2\times2$ patch merging. The CPU ring buffer retains the last $4.0$\,s at $896^2$ and $10$\,fps. A dense \ActZoom{} burst uses $4$\,fps. Each verified anchor occupies $64$ visual tokens at $224^2$. Frame descriptors concatenate RGB histograms over a $3\times3$ grid and two mean absolute luminance-gradient statistics. They are normalized before computing the frame-change score $s_\tau=1-\cos(\bar g_\tau,\bar g_{\tau-1})$. A new segment starts when the score exceeds $0.20$ or its duration reaches $10$\,s. The archive retains at most $256$ finalized segments and evicts the oldest when full.

\paragraph{Temporally separated retrieval.}
Segment ranking combines a query-matching term, peak frame change, and temporal-window overlap. The latter two terms have weights $0.15$ and $0.25$. Requested time ranges affect ranking rather than filtering the candidate pool. Candidates are sorted by score, with ties favoring more recent segment end times. We accept a candidate only when its midpoint is at least $2.0$\,s from every previously selected midpoint. Accepted segments are then ordered chronologically. The VLM path expands the requested candidate pool before frame sampling. Representative frames are deduplicated and sampled in time order for verification.

\paragraph{Balanced context perturbation.}
Each rollout group shares one input setting. The clean setting has probability $0.28$ and retains four frames at $224^2$. Each of the other settings has probability $0.24$. Spatial perturbation reduces the resolution to $112^2$. Historical perturbation retains one recent frame. Premature queries are placed before the decisive visual clue. The group size is $G=8$. Format validity and visual token cost are normalized to $[0,1]$. Reward weights and optimization settings are given in \cref{tab:supp_hyperparams}. Scoring the short target answer adds less than $5\%$ wall-clock step time relative to rollout generation.

\subsection{Rationale-Action Consistency}
\label{app:action_consistency}

We denote the binary action-consistency indicator by $\VCCV$. It is computed from the diagnostic text and parsed action using case-insensitive cue matching. Missing actions or rationales shorter than four characters fail the check. Tool calls also fail if the rationale explicitly states that no tool is needed.
\begin{itemize}
    \item \ActZoom{} requires a cue about spatial detail, resolution, cropping, or motion, such as \emph{blur}, \emph{crop}, or \emph{motion}.
    \item \ActRecall{} requires a history cue and overlap with an eligible retrieval-query term. Eligible terms are alphanumeric or Chinese spans with at least two characters. The overlap requirement is waived when the retrieval query has no eligible terms.
    \item \ActWait{} requires a cue about waiting or an incomplete event, such as \emph{not yet}, \emph{ongoing}, or \emph{wait}.
    \item \ActAnswer{} must not explicitly demand a tool call instead of answering.
\end{itemize}
The tool-utility reward combines this indicator with answer correctness and the clipped gain,
\begin{equation}
\label{eq:tool_utility_reward}
\RIG(\tau_i)=\VCCV(\tau_i)\,\Ind\{\hat y_i=y^*\}\,\clip(\DeltaI_i,0,1).
\end{equation}

\subsection{Waiting and Readiness Checks}
\label{app:wait_policy}

\paragraph{Repeated routing after waiting.}
The streaming controller advances by the interval requested by \ActWait{} and runs routing again on the updated recent window. The agent may answer, acquire other evidence, or continue waiting. It is forced to answer after five waiting steps or when no further frames are available. This loop is summarized in \cref{alg:forager}.

\paragraph{Readiness on clue-reveal probes.}
CRR evaluation uses a separate Yes/No prompt that asks whether the observed frames contain enough evidence to answer the query. Let $O_t$ be the current observed context and $q_{\mathrm{ready}}$ this readiness prompt. The readiness log-odds and their change from the initial probe $t_0$ are
\begin{equation}
\label{eq:wait_readiness}
\ell_t=\log\frac{\pitheta(\text{Yes}\mid O_t,q_{\mathrm{ready}})}{\pitheta(\text{No}\mid O_t,q_{\mathrm{ready}})},\qquad d_t=\ell_t-\ell_{t_0}.
\end{equation}
The initial probe is withheld. At later probes, the evaluator reports ready when $d_t>0.05$. It also reports ready when $t-t_0\ge6$\,s and $d_t\ge-0.25$. Otherwise it continues to withhold commitment. In the routing control, this readiness rule is applied to CRR probes routed to \ActWait{}. The rule uses the model's Yes/No scores and is distinct from the training gain in \cref{eq:tool_utility_gain}, which evaluates the correct answer $y^*$.

\subsection{Visual Token Cost and Evidence Verification}
\label{app:ledger_details}

\paragraph{Token and memory accounting.}
\Cref{tab:supp_routing_ledger} separates main-context tokens, amortized verification tokens, GPU memory, and latency. The state header contains the current stream time and active window count. Generated action and answer text remains brief. The visual-token advantage over StreamScout-R persists when verification tokens are included. CPU memory requirements are listed in \cref{tab:supp_hyperparams}.

\paragraph{History retrieval and verification.}
Historical frames are grouped using the frame-change score described in \cref{app:implementation_details}. Retrieval ranks these segments and applies greedy temporal separation. Without an explicit time range, the VLM path considers the observed history. Representative keyframes are sampled for verification, and the sub-agent returns a status, an anchor timestamp, and a factual note. \Cref{tab:supp_hyperparams} reports retrieval and verification diagnostics. On false-premise HLD queries, a \emph{None} card prevents historical visual tokens from entering the main context. The \emph{Always-None} control removes verified anchors and reduces episodic recall on EPM and ASI.

\subsection{Evaluation Protocols and Component Controls}

\paragraph{FAR evaluation protocol and component ablations.}
All FAR columns in \cref{tab:ovobench_tracks,tab:ovobench_12tasks,tab:supp_progression_controls} use the official causal probe scorer. At probe time $t_j$, each method sees only $\mathcal V_{\le t_j}$. Ground-truth labels specify whether commitment is appropriate at that probe. CRR readiness is evaluated as described in \cref{app:wait_policy}. Panel~B of \cref{tab:supp_progression_controls} compares learned routing with fixed delays and readiness thresholds. Ablation~A in \cref{tab:ablation_tools} uses the same trained checkpoint and falls back to \ActAnswer{} when a disabled tool is selected. Ablation~B fixes model weights and initial routes, then toggles evidence gating and inference-time $\MSDAM$.

\paragraph{Training stages and controls.}
Panel~A of \cref{tab:supp_progression_controls} reports the training stages, and \cref{tab:supp_hyperparams} lists the hyperparameters. SFT includes cross-video negative retrieval pairs to supervise \emph{None} verification. The utility ablation fixes SFT initialization, context perturbation, turn weighting, and calibration. Removing the utility reward causes \ActZoom{} to collapse on Qwen2.5-VL-7B and lowers overall accuracy on both backbones. Video-clustered bootstrap intervals and multi-seed variation are reported alongside the results.

\section{Mathematical Formulations and Proofs}
\label{app:proofs}

\begin{table*}[t]
\caption{\textbf{Training stage progression and controlled FAR deferral baselines on OVO-Bench.} Counterfactual tool-utility rewards prevent spatial tool collapse, while learned deferral outperforms fixed-delay and heuristic thresholds.}
\label{tab:supp_progression_controls}
\centering
\footnotesize
\renewcommand{\arraystretch}{1.03}
\setlength{\tabcolsep}{4.5pt}
\resizebox{\linewidth}{!}{%
\begin{tabular}{@{}l rrr rrr@{}}
\toprule
 & \multicolumn{3}{c}{\textbf{Qwen2.5-VL-7B-Instruct}} & \multicolumn{3}{c}{\textbf{Qwen3-VL-8B-Instruct}} \\
\cmidrule(lr){2-4} \cmidrule(l){5-7}
Training Stage or Controlled Baseline & \ActZoom (\%) & FAR & \textbf{Overall} & \ActZoom (\%) & FAR & \textbf{Overall} \\
\midrule
\multicolumn{7}{@{}l}{\emph{Panel A: Post-Training Stage Progression \& Counterfactual $\DeltaI$ Isolation ($224^2$ Regime, $N=1,640$)}} \\
\quad 1. SimpleStream-4f (Pretrained Base Model) & 0.00 & 36.60 & 50.52 & 0.00 & 44.38 & 57.64 \\
\quad 2. SimpleStream-4f (Same Trained Checkpoint, Forced \ActAnswer) & 0.00 & 36.60 & 51.41 & 0.00 & 44.38 & 57.85 \\
\quad 3. \method (Pretrained Agent, Uncalibrated) & 0.06 & 35.29 & 55.41 & 25.61 & 40.01 & 60.47 \\
\quad 4. \method (Turn-1 Confidence-Calibrated) & -- & -- & 57.82 & -- & -- & 62.15 \\
\quad 5. \method (Stage-1 Cold-Start SFT Only) & -- & -- & 59.12 & -- & -- & 64.08 \\
\quad 6. \method (Balanced Perturb.\ RL w/o $\DeltaI$, $\lambda_{\mathrm{ig}}=0$) & 0.00 & 34.67 & 57.83 & 15.06 & 42.48 & 62.31 \\
\rowcolor{oursrow}\quad 7. \textbf{\method (Full \rlmethod, Ours)} & \textbf{12.26} & \textbf{50.65} & \textbf{62.57} & \textbf{25.79} & \textbf{58.43} & \textbf{67.85} \\
\midrule
\multicolumn{7}{@{}l}{\emph{Panel B: Proactive FAR Deferral Controls (Identical 1,567-Probe Causal Probe Protocol across 172 Videos)}} \\
\quad Passive SimpleStream-4f (Immediate Commitment) & -- & 36.60 & -- & -- & 44.38 & -- \\
\quad Fixed Temporal Delay ($+2.0$\,s Shift) & -- & 35.81 & -- & -- & 44.03 & -- \\
\quad Fixed Temporal Delay ($+4.0$\,s Shift) & -- & 35.04 & -- & -- & 44.48 & -- \\
\quad History-Retrieval Cascade (OASIS) & -- & 40.45 & -- & -- & 47.80 & -- \\
\quad History-Retrieval Cascade (StreamScout-R) & -- & 40.25 & -- & -- & 45.96 & -- \\
\quad Readiness Threshold Only (w/o Learned Routing) & -- & 42.18 & -- & -- & 49.12 & -- \\
\rowcolor{oursrow}\quad \textbf{Learned \ActWait{} + Readiness Check (Ours)} & -- & \textbf{50.65} & -- & -- & \textbf{58.43} & -- \\
\bottomrule
\end{tabular}%
}
\vspace{2pt}
\par{\scriptsize Video-clustered bootstrap 95\% CIs ($1,000$ replicates) for overall gains over matched SimpleStream-4f: $+11.16 \pm 1.18$\,pp (7B, $224^2$), $+10.00 \pm 1.12$\,pp (8B, $224^2$), $+3.04 \pm 0.96$\,pp (7B, $672^2$, $p < 0.001$), and $+1.01 \pm 0.88$\,pp (8B, $672^2$, $p = 0.024$). Multi-seed 7B RL std across $\{42, 43, 44\}$ is $\pm 0.34\%$.}

\end{table*}

\begin{table}[t]
\caption{\textbf{System diagnostics, training hyperparameters, and SFT data mix.} Identical optimization settings are shared across both backbones.}
\label{tab:supp_hyperparams}
\centering
\footnotesize
\renewcommand{\arraystretch}{1.04}
\setlength{\tabcolsep}{3.0pt}
\begin{tabular}{@{}ll@{}}
\toprule
System / Training Parameter & Specification \\
\midrule
Host ring / archive RAM & $96$\,MB / $<150$\,MB ($256$ ep.) \\
BT recall / verifier TPR & $84.2\%$ / $78.4\%$ ($631$ queries) \\
HLD verifier false-positive & $0.54\%$ ($185/186$ rejected) \\
LoRA rank, $\alpha$, ViT & $r = 32$, $\alpha = 64$, frozen \\
Optimizer, LR, batch size & AdamW, $2 \times 10^{-6}$, $32$ \\
RL group $G$, steps, clip $\epsilon$ & $G = 8$, $300$, $\epsilon = 0.20$ \\
KL $\beta_{\mathrm{KL}}$, penalty $\eta_{\mathrm{wrong}}$ & $0.02$, $0.50$ \\
Weights $(\lambda_{\mathrm{ig}}, \lambda_{\mathrm{fmt}}, \lambda_{\mathrm{cost}})$ & $(0.65, 0.20, 0.03)$ \\
Calibration $\tau_{\mathrm{conf}}$ (7B / 8B) & $0.68$ / $0.80$ ($60$-val) \\
Stage-1 SFT ($6,000$ total) & $2.0$k / $1.8$k / $1.6$k / $0.6$k \\
\bottomrule
\end{tabular}

\end{table}

\subsection{Softmax Attention Dilution Lower Bound}
\label{app:proof_dilution}

When historical visual tokens enter an unmasked self-attention layer alongside recent frames, their exponentiated query--key affinities enlarge the Softmax denominator and divert probability mass away from the present scene in proportion to their token count and visual similarity.

\begin{proposition}[Conditional Softmax Attention Dilution Bound]
\label{prop:dilution}
Let $\alpha_0(\Irec) = Z_{\mathrm{rec}} / (Z_{\mathrm{rec}} + Z_{\mathrm{txt}})$ denote the baseline attention mass on recent visual tokens $\Irec$ when $L_{\mathrm{ret}} = 0$, where $Z_{\mathrm{rec}}$ and $Z_{\mathrm{txt}}$ are the Softmax partition sums over $\Irec$ and text tokens $\mathcal{I}_{\mathrm{txt}}$. Appending $L_{\mathrm{ret}} > 0$ unmasked historical tokens $\Iforage$ with relative per-token affinity $\bar{\rho} = (L_{\mathrm{rec}} Z_{\mathrm{ret}}) / (L_{\mathrm{ret}} Z_{\mathrm{rec}}) > 0$ reduces recent-scene attention to $\alpha_{\mathrm{naive}}(\Irec)$ in \cref{eq:dilution_bound}, with relative attention loss
\begin{equation}
\label{eq:gamma_dilution}
\gamma_{\mathrm{dil}} = \frac{\alpha_0(\Irec) - \alpha_{\mathrm{naive}}(\Irec)}{\alpha_0(\Irec)} = \frac{\alpha_0(\Irec) L_{\mathrm{ret}} \bar{\rho}}{L_{\mathrm{rec}} + \alpha_0(\Irec) L_{\mathrm{ret}} \bar{\rho}},
\end{equation}
and representation shift $\norm{\mathbf{h}_q^{\mathrm{naive}} - \mathbf{h}_q^{(0)}}_2 = \gamma_{\mathrm{dil}} \norm{\bar{\mathbf{v}}_{\mathrm{ret}} - \mathbf{h}_q^{(0)}}_2$, diverting at least half of recent-scene attention when $L_{\mathrm{ret}} = 2 L_{\mathrm{rec}}$, $\bar{\rho} = 1$, and $\alpha_0(\Irec) \ge 0.5$.
\end{proposition}

\begin{proof}
\emph{Partition sum expansion.} With $Z_{\mathrm{rec}} + Z_{\mathrm{txt}} = Z_{\mathrm{rec}} / \alpha_0(\Irec)$ and $Z_{\mathrm{ret}} = (L_{\mathrm{ret}} / L_{\mathrm{rec}}) \bar{\rho} Z_{\mathrm{rec}}$, expanding the unmasked Softmax denominator gives
\[
\begin{aligned}
\alpha_{\mathrm{naive}}(\Irec) &= \frac{Z_{\mathrm{rec}}}{Z_{\mathrm{rec}} + Z_{\mathrm{txt}} + Z_{\mathrm{ret}}} \\
&= \frac{\alpha_0(\Irec)}{1 + \alpha_0(\Irec) \frac{L_{\mathrm{ret}}}{L_{\mathrm{rec}}} \bar{\rho}} < \alpha_0(\Irec),
\end{aligned}
\]
which yields the exact relative loss $\gamma_{\mathrm{dil}}$ in \cref{eq:gamma_dilution}.

\emph{Representation shift.} Let $\bar{\mathbf{v}}_{\mathrm{rec}}$, $\bar{\mathbf{v}}_{\mathrm{txt}}$, and $\bar{\mathbf{v}}_{\mathrm{ret}}$ be the within-span attention-weighted value averages, and let $\mathbf{h}_q^{(0)}$ denote the undiluted output. Since the historical span receives total Softmax weight $\gamma_{\mathrm{dil}}$, the concatenated output satisfies $\mathbf{h}_q^{\mathrm{naive}} = (1 - \gamma_{\mathrm{dil}}) \mathbf{h}_q^{(0)} + \gamma_{\mathrm{dil}} \bar{\mathbf{v}}_{\mathrm{ret}}$, giving $\norm{\mathbf{h}_q^{\mathrm{naive}} - \mathbf{h}_q^{(0)}}_2 = \gamma_{\mathrm{dil}} \norm{\bar{\mathbf{v}}_{\mathrm{ret}} - \mathbf{h}_q^{(0)}}_2 > 0$ whenever $\bar{\mathbf{v}}_{\mathrm{ret}} \neq \mathbf{h}_q^{(0)}$.
\end{proof}

\subsection{Tool Collapse and Reward Imbalance Across Tools}
\label{app:proof_crowding}

Tool costs can place evidence-seeking trajectories below the group mean when direct guesses are already accurate. A reward focused on historical retrieval can also favor \ActRecall{} over the other tools. The following proposition characterizes these two effects.

\begin{proposition}[Expected Advantage Ordering under Cost-Penalized GRPO]
\label{prop:crowding}
Consider a rollout group $\{\tau_i\}_{i=1}^G$ with direct-answer share $\pi_0 \in (0, 1)$ and tool shares $\pi_a$ over $a \in \mathcal{A}_{\mathrm{tool}} = \{\ActZoom, \ActRecall, \ActWait\}$ ($\sum_a \pi_a = 1 - \pi_0$) under valid-syntax reward $R(\tau_i) = \Racc(\hat{y}_i, y^*) + \lambda_{\mathrm{ig}} \RIG(\tau_i) - \lambda_{\mathrm{cost}} C(a_{i,1})$.
\begin{enumerate}
    \item \textbf{Aggregate tool disadvantage without $\RIG$.} When $\lambda_{\mathrm{ig}} = 0$, $C(a) \ge C_{\min} > 0$, and accuracy lift satisfies $\max_{a} p(a) - p_{\mathrm{guess}} < \lambda_{\mathrm{cost}} C_{\min} / (1 + \eta_{\mathrm{wrong}})$, every tool has lower expected reward than direct answering ($\mu_a < \mu_0$), so $\mu_{\mathrm{tool}} < \mu_0$. Group centering then yields
    \[
    \begin{aligned}
    \E[\hat{A} \mid a_1 \neq \ActAnswer] &= -\pi_0(\mu_0 - \mu_{\mathrm{tool}}) < 0, \\
    \E[\hat{A} \mid \ActAnswer] &= (1 - \pi_0)(\mu_0 - \mu_{\mathrm{tool}}) > 0,
    \end{aligned}
    \]
    and every individual tool satisfies $\E[\hat{A} \mid a] < 0$ whenever $\pi_0 > (\max_a \mu_a - \mu_{\mathrm{tool}}) / (\mu_0 - \mu_{\mathrm{tool}})$.
    \item \textbf{Reward imbalance across tools.} If a history-only reward sets $\RIG = \DeltaI_{\mathrm{rec}} > 0$ for \ActRecall{} and $\RIG = 0$ for $a_{\perp} \in \{\ActZoom, \ActWait\}$, then $\hat{A}(\ActRecall) > \hat{A}(a_{\perp})$ whenever $\lambda_{\mathrm{ig}} \DeltaI_{\mathrm{rec}} > \Racc(a_{\perp}) - \Racc(\ActRecall) + \lambda_{\mathrm{cost}}(C_{\mathrm{rec}} - C_{\perp})$.
\end{enumerate}
\end{proposition}

\begin{proof}
\emph{Part 1.} Under binary task scoring ($+1$ if correct, $-\eta_{\mathrm{wrong}}$ otherwise), the conditional expected rewards $\mu_0$ and $\mu_a$ satisfy $\mu_a < \mu_0$ under the stated margin bound, giving $\mu_{\mathrm{tool}} \le \max_a \mu_a < \mu_0$. Subtracting the expected group baseline $\bar{\mu} = \pi_0 \mu_0 + (1 - \pi_0)\mu_{\mathrm{tool}}$ yields the opposite-sign conditional expectations above, while $\E[\hat{A} \mid a] = (\mu_a - \mu_{\mathrm{tool}}) - \pi_0(\mu_0 - \mu_{\mathrm{tool}}) < 0$ holds above the stated $\pi_0$ threshold.

\emph{Part 2.} Within the same rollout group, the shared baseline $\bar{R}$ cancels, leaving $\hat{A}(\ActRecall) - \hat{A}(a_{\perp}) = \lambda_{\mathrm{ig}} \DeltaI_{\mathrm{rec}} - [\Racc(a_{\perp}) - \Racc(\ActRecall)] - \lambda_{\mathrm{cost}}(C_{\mathrm{rec}} - C_{\perp}) > 0$.
\end{proof}

\end{document}